\documentclass[]{template/mac_automl_style}
\makeatletter
\def\input@path{{content/}}
\makeatother
\graphicspath{{content/}}

\usepackage[toc,page,header]{appendix}

\usepackage{graphicx}
\usepackage{booktabs}
\usepackage{multirow}
\usepackage{pifont}
\usepackage{amssymb}
\usepackage{amsfonts}
\usepackage{amsmath}
\usepackage{algorithm}
\usepackage{algpseudocode}
\usepackage{colortbl}
\usepackage{makecell}
\usepackage{enumitem}
\usepackage{xcolor}
\usepackage{xspace}
\usepackage{gradient-text}
\usepackage{tabularx}
\usepackage{array}
\usepackage{diagbox}
\usepackage{nicefrac}
\usepackage{wrapfig}
\usepackage{bbm}
\usepackage{listings}
\usepackage{fontawesome5}
\usepackage{titletoc}
\usepackage{amsthm}
\algrenewcommand{\algorithmicrequire}{\textbf{Input:}}
\algrenewcommand{\algorithmicensure}{\textbf{Output:}}
\algnewcommand{\REQUIRE}{\Require}
\algnewcommand{\ENSURE}{\Ensure}
\algnewcommand{\STATE}{\State}
\algnewcommand{\FOR}{\For}
\algnewcommand{\ENDFOR}{\EndFor}
\algnewcommand{\IF}{\If}
\algnewcommand{\ELSE}{\Else}
\algnewcommand{\ENDIF}{\EndIf}
\algnewcommand{\COMMENT}[1]{\Comment{#1}}

\tcbset{
  aibox/.style={
    width=\linewidth,
    top=7pt,
    bottom=2pt,
    colback=blue!4!white,
    colframe=black,
    colbacktitle=black,
    enhanced,
    center,
    attach boxed title to top left={yshift=-0.1in,xshift=0.15in},
    boxed title style={boxrule=0pt,colframe=white,},
  }
}
\newtcolorbox{AIbox}[2][]{aibox,title=#2,#1}

\newcommand{\ours}{\gradientRGB{YOLO-PEFT}{29,78,216}{20,184,166}\xspace}
\newcommand{\system}{\ours}

\newcommand{\modB}{Planner\xspace}
\newcommand{\modC}{Contract\xspace}

\definecolor{my_green}{RGB}{34, 139, 34}
\definecolor{my_red}{RGB}{220, 20, 60}
\definecolor{my_orange}{RGB}{255, 140, 0}
\definecolor{my_lightgray}{RGB}{245, 245, 245}
\definecolor{my_darkgray}{RGB}{100, 100, 100}

\newcommand{\authormark}[1]{\textsuperscript{#1}}

\newcommand{\authorentry}[2]{\mbox{#1\authormark{#2}}}
\newcommand{\authorsep}{\hspace{0.78em}}

\setleftheadercontent{%
  \headerlogospace{2.4mm}%
   \adjustbox{valign=c}{\raisebox{0.2mm}{\includegraphics[height=11.5mm]
    {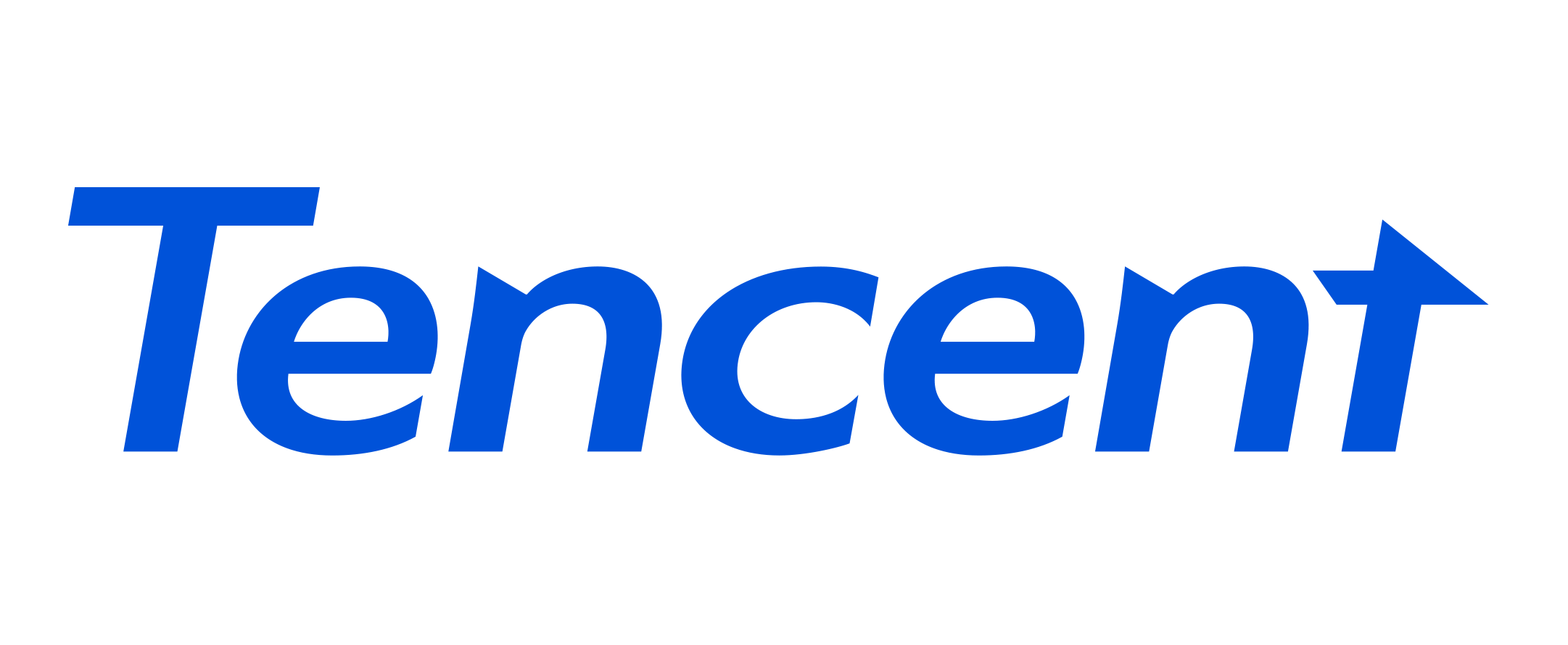}}}%
  \headerlogospace{1.6mm}%
  \adjustbox{valign=c}{\raisebox{0.15mm}{\includegraphics[height=15.2mm]{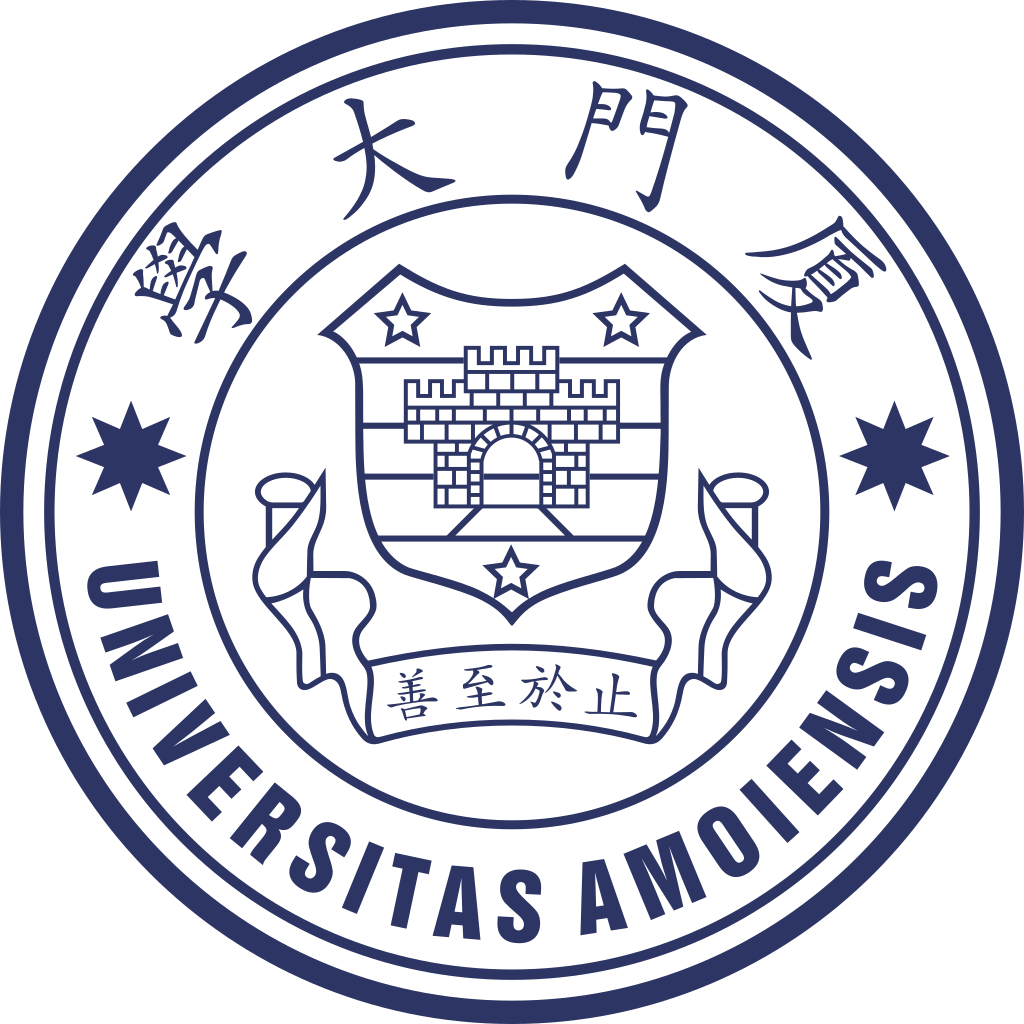}}}%
}
\setrightheadericon{%
  \adjustbox{valign=c}{\raisebox{0.55mm}{\includegraphics[height=14.8mm]{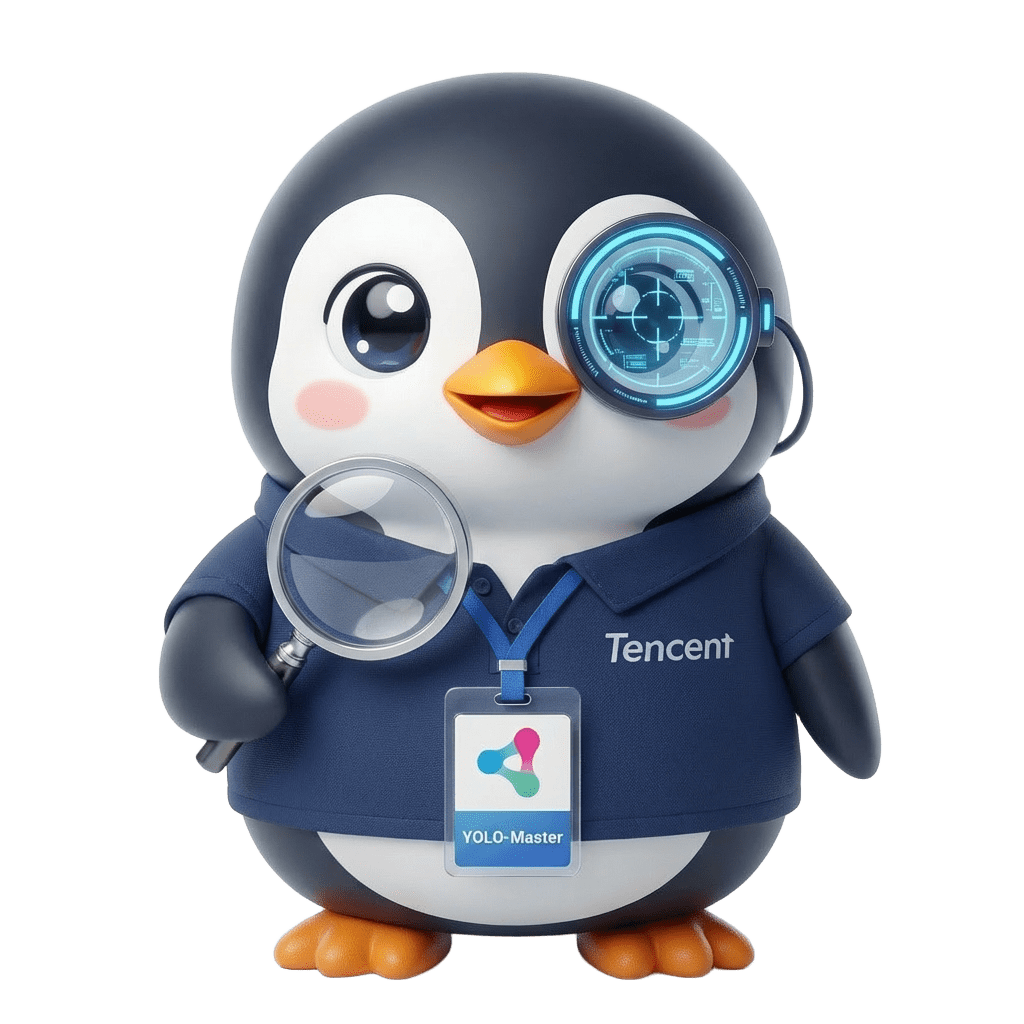}}}%
}

\setrunningheadericon{%
  
}
\setheadergroupname{}

\hypersetup{
  breaklinks=true,
  colorlinks=true,
  citecolor=omnilink,
  linkcolor=omnilink,
  urlcolor=omnilink
}

\title{\ours: Parameter-Efficient Fine-Tuning on YOLO Family}

\makeatletter
\def\authorlist{%
  {
  \fontsize{12}{14}\selectfont\color{black!96}%
  \ifXeTeX\omniauthorfont\bfseries\else\sffamily\bfseries\fi
  \parbox{0.96\linewidth}
    {\centering
    \makebox[\linewidth][c]{%
      \authorentry{Xu Lin}{1},\authorsep
      \authorentry{WenJie Nie}{2},\authorsep
      \authorentry{Jinlong Peng}{1},\authorsep
      \authorentry{Weifu Fu}{1},\authorsep
      \authorentry{YueXiao Ma}{2},\authorsep
      \authorentry{Xiawu Zheng}{2},\authorsep
      \authorentry{Yong Liu}{1}
      }%
    }%
  }
}
\def\affiliationlist{%
  {\fontsize{9.9}{11.9}\selectfont\color{black!70}%
  \ifXeTeX\omnibytesansmedium\else\sffamily\fi
  \parbox{0.92\linewidth}{\centering
  \authormark{1}Tencent
  \authormark{2}Xiamen University\par
  \vspace{2.8mm}
  {\fontsize{9.8}{11.8}\selectfont\color{black!82}%
  \ifXeTeX\omnibytesansmedium\else\sffamily\fi
  {  
  }}}}}
\def\emaillist{%
  {}}
\makeatother

\abstract{Generic parameter-efficient fine-tuning (PEFT) methods transferred from language
models can fail silently on real-time detectors, whose heterogeneous operators and
detection-specific components impose placement constraints absent from regular
Transformer stacks. We propose YOLO-PEFT, a structure-aware framework that formulates
adapter placement as an auditable constraint-planning problem. Given a detector graph,
a PEFT request, and a resource budget, YOLO-PEFT assigns operator and semantic roles,
evaluates explicit operator-validity, detector-semantic, graph-interface, and deployment
predicates, records a reason code for each excluded module, and either emits a budgeted
target-module plan or returns Refuse before training. Under the official VOC07+12
trainval-to-VOC07 test protocol, planner-selected RS-LoRA reaches 0.7138 and 0.7307
mAP50-95 on YOLO11s and YOLO12s, respectively, compared with 0.6428 and 0.6662 for
Full-SFT. On RT-DETR-L, all seven evaluated LoRA-family configurations cross the
predefined catastrophic threshold, supporting a calibrated Refuse-to-Full-SFT decision
within the evaluated coverage. A controlled YOLO11 audit further shows that LoRA
reduces peak training memory by 43.9 percent, although training takes 1.72 times
longer. Within the evaluated detector families, placement policies, and calibration
coverage, YOLO-PEFT replaces manual target-module trial and error with explicit,
inspectable planning while preserving verified train-save-merge-export paths; refusal
on unseen detector architectures remains an open validation problem.
}
\checkdata[Project Page]{\href{https://github.com/Tencent/YOLO-Master}{\nolinkurl{github.com/Tencent/YOLO-Master}}}

\begin{document}
\maketitle

\section{Introduction}
\label{sec:intro}

Deploying real-time detectors at scale requires recurring adaptation to new class
vocabularies, domains, sensors, latency targets, and hardware backends. This work studies
representative members of the broader YOLO ecosystem together with RT-DETR, YOLO-World,
and a mixture-of-experts (MoE) detector~\cite{cheng2024yoloworld,jocher2024yolo11,lv2024rtdetr,wang2025yolov12}.
Full fine-tuning replicates training state and complete checkpoints for every deployment
variant, making parameter-efficient fine-tuning (PEFT) an attractive way to reduce
adaptation and distribution cost~\cite{han2024peftsurvey,he2022towards,hu2022lora}.
The open challenge is that PEFT developed for largely homogeneous Transformer stacks does
not transfer safely to heterogeneous detector graphs: modern detectors interleave dense,
grouped, and depthwise convolutions, loss-coupled Distribution Focal Loss (DFL)
projections, deformable attention, text--image fusion, and MoE routing, so indiscriminate
LoRA insertion can cause optimization failure or severe mAP degradation.

Existing solutions leave four gaps. First, generic PEFT interfaces select targets by
module name or type~\cite{mangrulkar2022peft}, without checking operator contracts,
detection semantics, graph interfaces, or placement risk before training. Second,
detector-specific methods such as SpotPatch, LoRA-Det, and YOLO-IOD demonstrate the value
of selective adaptation, but their placement policies are tied to particular detectors,
tasks, or components rather than a common abstraction spanning convolutional,
Transformer, multimodal, and MoE graphs~\cite{pu2025loradet,ye2021spotpatch,zhang2026yoloiod}.
Third, placement alone does not provide an end-to-end contract for training, adapter-only
checkpointing, merging, and ONNX/TensorRT export, leaving deployment compatibility to
method-specific integration. Fourth, prior interfaces provide no calibrated mechanism
for identifying high-risk architecture--adapter combinations before committing to a full
training run; failure is therefore discovered mainly through costly trial and error.

\begin{figure*}[!t]
\centering
\includegraphics[width=0.95\textwidth]{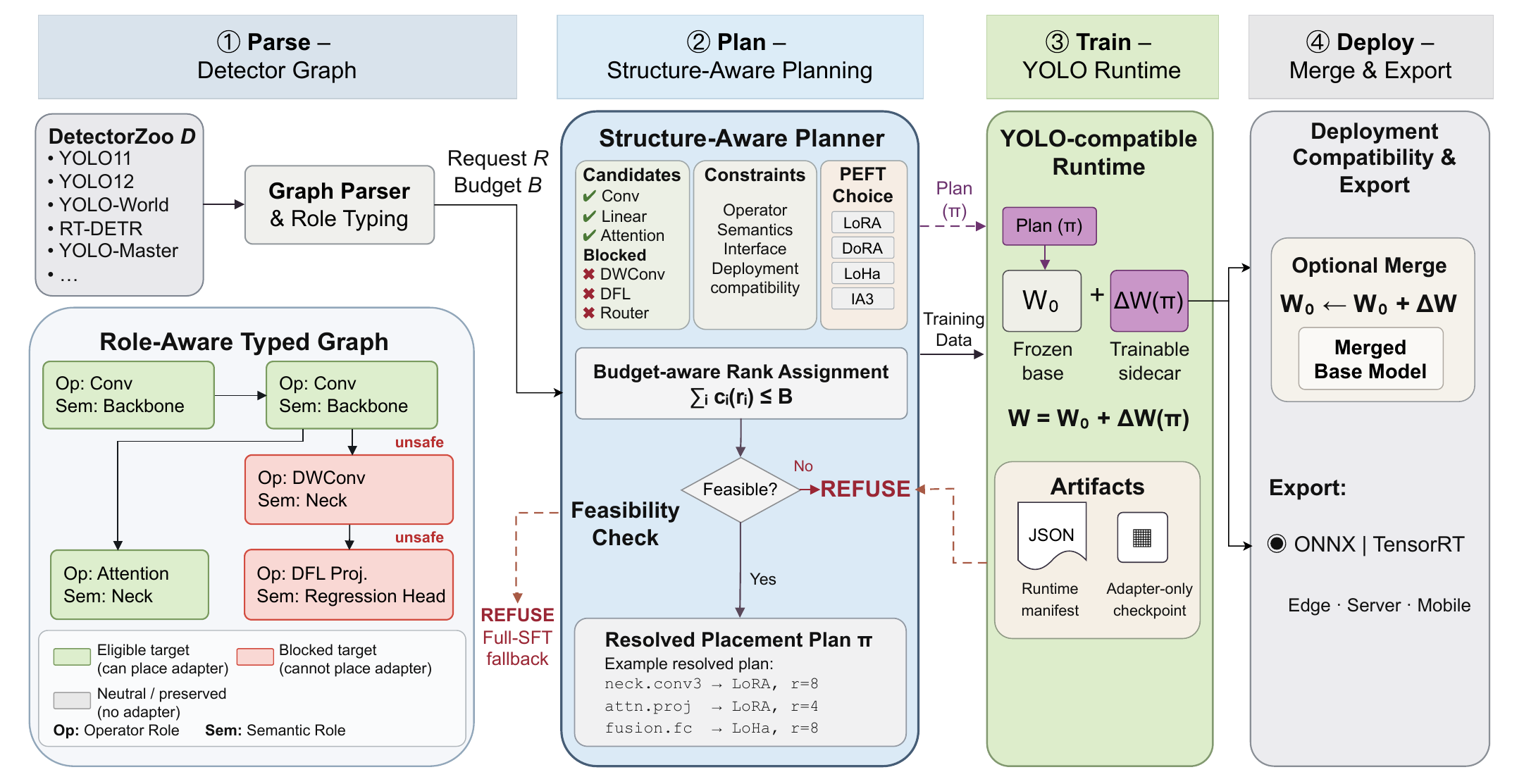}

% \caption{\textbf{Training-time GPU memory.} YOLO11 and YOLO12
% under LoRA with $r{=}8$, gradient checkpointing on. Our
% reduction on every scale.\label{fig:overview}}
% \caption{\textbf{System overview.} The pipeline
% (1)~parses a YOLO-family detector into a typed graph,
% (2)~resolves adapter placements~$\pi$ via a constraint gate under a
% budgeted rank assignment $\sum_i c_i(r_i)\!\le\!B$,
% (3)~trains sidecar adapters with base weights frozen
% ($W{=}W_0{+}\Delta W$), and
% (4)~exports either an adapter-only checkpoint or a merged, ordinary
% YOLO model (ONNX/TensorRT).\label{fig:overview}}
% \end{figure*}
% \caption{\textbf{System overview.} The pipeline
% (1)~parses a YOLO-family detector into a typed graph,
% (2)~resolves adapter placements~$\pi$ via a constraint gate under a

% deployment.\label{fig:overview}}
\caption{\textbf{System overview.} The pipeline
\textbf{(1)}~parses a detector from a supported, evaluated family into a role-aware typed graph,
\textbf{(2)}~filters unsafe targets, selects the PEFT method, and resolves an adapter
placement~$\pi$ via budget-aware rank assignment
$\sum_i c_i(r_i)\!\le\!B$, or, within calibrated coverage, returns \textsc{Refuse} with a Full-SFT fallback,
\textbf{(3)}~trains sidecar adapters while freezing the base weights
($W{=}W_0{+}\Delta W(\pi)$) and saves an adapter-only checkpoint with its
runtime manifest, and
\textbf{(4)}~optionally merges the adapters into an ordinary YOLO model and exports
the resulting model through ONNX/TensorRT for edge, server, or mobile
deployment.\label{fig:overview}}
\end{figure*}

Rather than proposing another low-rank parameterization, we recast detector PEFT as
multi-constraint planning on a heterogeneous computation graph and introduce YOLO-PEFT,
a structure-aware framework that parses operator and semantic roles, filters unsafe
targets, allocates ranks under a resource budget, estimates failure risk within calibrated
coverage, returns \textsc{Refuse} when no reliable plan exists, and lowers accepted plans
to a unified train--save--merge--export runtime (Fig.~\ref{fig:overview}).

Our contributions follow the progression from formulation to evidence:
\begin{itemize}
\item \textbf{Constrained formulation and safety model.} We define adapter placement by
joint operator-validity, detection-semantic, graph-interface, deployment, and budget
constraints. A dual-role graph parser and a bounded 10-dimensional architecture
fingerprint characterize detector structure; its calibrated core dimensions support
auditable pre-training risk estimation within the evaluated calibration coverage, while
\textsc{Refuse} makes fallback to Full-SFT a first-class planning outcome rather than a
runtime failure. This is not a safety guarantee for an unseen detector family.

\item \textbf{Constraint-resolved planning.} We develop an ordered, multi-level planner
that enforces mandatory operator and semantic filters, applies detector-specific safeguards
for DFL paths, attention modules, text fusion, and MoE routing, and solves budget-aware
rank assignment over the surviving targets.

\item \textbf{Deployment-complete runtime contract.} We provide a YOLO-compatible
lifecycle for training, adapter-only persistence, reloading, weight merging, and edge
export through ONNX/TensorRT. We further establish merge equivalence for the fallback
grouped-convolution LoRA backend, showing that wrapper removal preserves convolution
outputs up to numerical tolerance.

\item \textbf{Systematic architecture-conditioned evidence within stated coverage.}
A diagnostic sweep spanning 14 PEFT variants and five detector architectures, plus a
separately scoped seed-0 YOLO-Master-EsMoE-S stress test with planner/expert targeting
disabled, shows no universal PEFT ordering over the evaluated families. Under the VOC
protocol, planner-selected LoRA outperforms Full-SFT on YOLO11s and YOLO12s ($+7.1$ and
$+6.5$ mAP$_{50:95}$); RT-DETR-L instead returns \textsc{Refuse}. The controlled YOLO11
audit measures $43.9\%$ lower peak VRAM but $1.72\times$ longer training. We do not claim
held-out-family safety, COCO/domain-shift generalization, or expert-aware MoE planning.

% \item \textbf{Systematic architecture-conditioned evidence.} 

\end{itemize}

% Overall, YOLO-PEFT turns PEFT on YOLO from a manual target-selection heuristic into a structure-constrained planning and deployment problem, enabling adapter placement that is valid, compact, and deployable across heterogeneous YOLO-family detectors.

% \begin{figure}[!t]
% \centering
% \includegraphics[width=0.80\linewidth]{Figs/fig1-v260601-overview.png}
% \caption{\textbf{Training-time GPU memory.} YOLO11 and YOLO12
% under LoRA with $r{=}8$, gradient checkpointing on. Our
% reduction on every scale.\label{fig:overview}}
% \end{figure}

% ---- Inlined source: Sec/5_related_work.tex ----
% Sec/5_related_work.tex
% Related Work for YOLO-PEFT

\section{Related Work}
\label{sec:related}

\subsubsection{Parameter-Efficient Fine-Tuning.}
PEFT freezes most pretrained weights and learns compact task-specific parameters. LoRA represents weight updates with low-rank factors~\citep{hu2022lora}; later methods refine the parameterization, allocation, or geometry, including IA$^3$~\citep{liu2022ia3}, AdaLoRA~\citep{zhang2023adalora}, DoRA~\citep{liu2024dora}, and HRA~\citep{yuan2024hra}. Unified libraries expose these methods through module-name or module-type targets~\citep{mangrulkar2022peft}, while AutoPEFT searches adapter configurations in pretrained language models~\citep{zhou2024autopeft}. Placement and allocation are also active research questions: KAdaptation selects vision-Transformer submodules using local intrinsic dimension~\citep{he2023visionadaptation}, sensitivity-aware tuning allocates a fixed budget to task-important weights or modules~\citep{he2023spt}, and salient-channel tuning selects task-dependent feature channels~\citep{zhao2024sct}. These methods make target selection data-dependent, but operate over regular backbone parameterizations and statistical importance. Our concern is complementary: whether a proposed host is admissible under detector-graph interfaces and semantics, and whether the complete plan remains budget-feasible and deployable.

\subsubsection{Vision and Dense-Prediction PEFT.}
Vision-side PEFT includes residual adapters~\citep{rebuffi2017learning}, visual prompt tuning~\citep{jia2022vpt}, feature scaling and shifting~\citep{lian2022scaling}, AdaptFormer~\citep{chen2022adaptformer}, convolutional bypasses for vision Transformers~\citep{jie2024convpass}, and Conv-Adapter for ConvNets~\citep{chen2024convadapter}. Pro-Tuning further supports both CNN and Transformer backbones and evaluates detection and segmentation~\citep{nie2024protuning}. Dense-prediction studies move beyond classification: LoRand inserts low-rank synthesized adapters into Swin blocks for detection and segmentation~\citep{yin2023lorand}; VMT-Adapter shares and refines features across multiple dense tasks~\citep{xin2024vmtadapter}; and VFM-Adapter combines dynamic local and global operations for foundation-model detection and segmentation~\citep{chen2025vfmadapter}. E$^3$VA shows that few trainable parameters alone do not guarantee low training memory or time, and constructs a separate gradient highway for Swin-based dense prediction~\citep{yin2024e3va}. A related preprint uses parallel multi-task adapters for camouflaged-object segmentation~\citep{xing2023paddetect}. This literature establishes that adapter structure, task, and systems cost all matter, but largely assumes a chosen backbone and head rather than validating a heterogeneous detector-wide target plan.

\subsubsection{Vision-Language and Open-Vocabulary Adaptation.}
VL-Adapter benchmarks parameter-efficient adapters for vision-language models~\citep{sung2022vladapter}, providing a foundation for multimodal adaptation. Within open-vocabulary detection, SIA-OVD learns shape-specific region adapters and their allocation to reduce the CLIP image--region gap~\citep{wang2024siaovd}, while modular PEFT studies task-specific component tuning in an open-vocabulary detector~\citep{faye2024lmpet}. PET-DINO~\citep{Fu_2026_CVPR} unifies visual cues in Grounding DINO for visual prompt detection. These approaches demonstrate that insertion and allocation should reflect region or modality semantics. They target a fixed vision-language pipeline, whereas \system{} additionally reasons about convolution groups, multi-scale neck interfaces, loss-coupled head components, expert routing, and export-time merging.

\subsubsection{Parameter-Efficient Detector Adaptation.}
Several studies directly adapt object detectors. SpotPatch learns dataset-specific gates over MobileNetV2--SSD-FPNLite layers and stores selected 1-bit residuals across ten detection tasks~\citep{ye2021spotpatch}. LoRA-Det analyzes rank for a Swin-based oriented detector and manually combines LoRA with full tuning in its FPN, RPN, and prediction components~\citep{pu2025loradet}, while multi-point insertion tuning places lightweight modules at several locations of a frozen small-object detector~\citep{goto2025mpit}. DA-Ada learns domain-aware adapters for unsupervised domain-adaptive detection~\citep{li2024daada}; CULoRA targets few-shot source-free adaptation~\citep{yao2025culora}; and CoPEFT combines a collaboration adapter with agent prompts for low-cost collaborative perception adaptation~\citep{wei2025copeft}. Most directly for this paper, YOLO-IOD performs stage-wise PEFT of YOLO-World by selecting task-important convolution kernels for incremental detection~\citep{zhang2026yoloiod}; SF-YOLO instead adapts YOLO through a teacher--student source-free scheme~\citep{varailhon2024sfyolo}. Thus, prior detector work has studied layer, insertion-point, component, and kernel selection. It does not, however, provide a common pre-training feasibility check over operator constraints, detection semantics, graph interfaces, and export behavior across heterogeneous detector families.

Recent readable preprints broaden this evidence. REAL-OW ablates LoRA rank and attention-projection placement in a rehearsal-free Deformable-DETR pipeline~\citep{zhang2026realow}. A cross-domain aerial study finds that direct LoRA on DiffusionDet can underperform full tuning, while applying it after intermediate full tuning is more competitive~\citep{talaoubrid2025aeriallora}. DroneFINE adds domain-aware hyper-adapters to Grounding DINO~\citep{wu2026dronefine}, and EW-DETR studies incremental low-rank adaptation of DETR~\citep{monga2026ewdetr}. LORS~\citep{li2024lors} cuts query-based decoder parameters via low-rank residual structures. These preprints are not treated as peer-reviewed evidence, but their architecture- and regime-dependent outcomes reinforce the need to validate a plan rather than assume a universal PEFT ranking.

\subsubsection{Heterogeneous Real-Time Detectors.}
Modern real-time detectors mix execution paradigms: RT-DETR adds a Transformer decoder~\citep{lv2024rtdetr}, YOLO-World adds text--image fusion~\citep{cheng2024yoloworld}, YOLOv12 introduces attention-centric blocks~\citep{wang2025yolov12}, and YOLO-Master~\citep{yolomaster2026} exemplifies sparse expert routing. Their heads contain loss-coupled components such as distributional box regression~\citep{li2020gfl}. This heterogeneity motivates the scope of \system{}: it does not propose another adapter parameterization, but validates and lowers existing ones into a common train--save--merge--export contract while preserving detector-specific interfaces.

% ---- Inlined source: Sec/2_method.tex ----
% Sec/2_method.tex
% Methodology for YOLO-PEFT

% \section{Methodology}
% \label{sec:method}

% \subsection{Problem Formulation}
% \label{sec:problem}

% % TODO: Formalize the adapter placement problem
% % Define detector graph, PEFT request, resource budget

% \subsection{Structure-Aware Graph Parser}
% \label{sec:parser}

% % TODO: Describe \modA (GraphParser)
% % - Operator role tagging (dense/grouped/depthwise conv, linear, attention)
% % - Semantic role tagging (backbone, neck, classification head, DFL projection, MoE router, etc.)

% \subsection{Constraint-Resolved Planner}
% \label{sec:planner}

% % TODO: Describe \modB (Planner)
% % - Operator validity filtering
% % - Semantic safety filtering
% % - Graph-interface compatibility checking
% % - Budget-aware rank assignment
% % - Refusal mechanism for infeasible plans

% \subsection{YOLO-Compatible Execution Contract}
% \label{sec:contract}

% % TODO: Describe \modC (Contract)
% % - Bridge generic PEFT wrappers with Ultralytics pipeline
% % - Feature routing, checkpointing, fusion safety
% % - Adapter-only save/load/merge/export
% % - ONNX/TensorRT export with zero inference overhead

% \subsection{Deployment Runtime}
% \label{sec:runtime}

% % TODO: Describe \modD (Manifest)
% % - Train--save--merge--export pipeline
% % - Adapter merging and model fusion
% % - Open-source runtime implementation

% Methodology v260715 by linxu
\section{Methodology}
\label{sec:method}

The evaluated detector families contain heterogeneous operators and task-specific dense heads, so
language-model PEFT rules cannot be transferred by matching module names alone.
YOLO-PEFT instead treats adapter placement as a constrained decision problem: given a
detector graph, a PEFT request, and a parameter budget, it returns a structurally valid,
trainable, and deployable plan, or refuses the request when no admissible plan exists
within the implemented rules and calibration coverage. The pipeline proceeds in four
stages: parse the detector, resolve a constraint-valid placement, lower the
plan into the YOLO runtime, and apply compatible training strategies. Complete
pseudocode and implementation details are provided in the supplementary materials.

% \subsection{Problem Formulation and Detector Representation}
% \label{sec:method:problem}
% \label{sec:method:parser}

% Let a detector $\mathcal{D}$ be a directed acyclic graph $G=(V,E)$, where $V$ contains
% module instances and $E$ contains tensor-flow edges. A PEFT request is
% $R=(p,T,L,K)$: variant $p$, optional target set $T$, optional layer interval $L$, and
% allowed ranks $K$. Given adapter budget $B$, the framework returns a placement $\pi$ or
% $\textsc{Refuse}$. For the surviving candidates $V_{\rm cand}$,
% \begin{equation}
% \pi:V_{\rm cand}\rightarrow\{0\}\cup K,\qquad
% \pi(i)=0\ \text{leaves layer }i\text{ frozen}.
% \end{equation}
\subsection{Problem Formulation and Detector Representation}
\label{sec:method:problem}

Let a detector $\mathcal{D}$ be represented as a directed acyclic graph
$G=(V,E)$, where $V$ contains module instances and $E$ contains
tensor-flow edges. A PEFT request is denoted by
$R=(p,T,L,K)$, consisting of a variant $p$, an optional target set $T$,
an optional layer interval $L$, and a set of allowed ranks $K$.
Given an adapter budget $B$, the framework returns either a placement
$\pi$ or \textsc{Refuse}. For the surviving candidate modules
$V_{\mathrm{cand}}$, the placement is defined as
\begin{equation}
    \pi \colon V_{\mathrm{cand}} \to \{0\} \cup K .
    \label{eq:adapter-placement}
\end{equation}
Here, $\pi(i)=0$ indicates that layer $i$ remains frozen, whereas
$\pi(i)=r\in K$ assigns a rank-$r$ adapter to that layer.
A valid plan must jointly satisfy operator validity, detection-head semantic safety,
graph-interface compatibility, parameter-budget feasibility, and deployment
compatibility. Refusal is therefore a legitimate solution when these constraints have
no feasible intersection or reliability calibration predicts catastrophic degradation.

To make this decision auditable, each candidate module $i$ is represented by
$z_i=(o_i,s_i,q_i,h_i)$: operator metadata $o_i$ (type, shape, groups, kernel and
initialisation), semantic role $s_i$ (backbone, neck, classification, regression, DFL,
decoder, router or expert), graph-interface metadata $q_i$ (input/output arity, tensor
shape and residual/fusion edges), and deployment metadata $h_i$ (backend support,
mergeability and exporter support). Each constraint returns a Boolean decision and a
reason code. The planner output is therefore
\begin{equation}
  \mathcal{P}(G,R,B)=(d,\pi,\mathcal{J}),\qquad
  d\in\{\textsc{Accept},\textsc{Refuse}\},
  \label{eq:planner-output}
\end{equation}
where $\mathcal{J}$ is the ordered rejection log. An accepted plan records excluded
modules as well as installed targets; a refusal returns an empty placement and at least
one terminal code. This distinguishes ``unsupported'' from ``predicted inaccurate'' and
prevents the same word \textsc{Refuse} from hiding different causes.

A module's Python type is insufficient to determine whether it is a safe target: the
same \texttt{Conv2d} may belong to a backbone block, feature-pyramid fusion, a regression
head, or a fixed DFL projection. \textsc{GraphParser} consequently assigns every module
two independent roles. The \emph{operator role} captures its computational contract.
Dense convolutions admit ordinary low-rank factors, grouped convolutions require
group-local factors, and depthwise convolutions cannot accept updates that mix channels.
Normalization, activation, and unknown operators are treated conservatively. The
\emph{semantic role} captures detector-specific meaning. Fixed DFL projections,
geometry-sensitive regression paths, and MoE routers are excluded because small updates
can alter calibrated bins or expert assignment even when their tensor shapes remain
valid.

Role assignment combines graph position with detector idioms. The parser recognises
YOLO12 Area-Attention blocks, RT-DETR deformable-attention modules, YOLO-World text
fusion, and MoE router/expert structures. Known families receive calibrated role
profiles, while custom detectors fall back to a topology scan; unmatched modules receive
an \texttt{unknown} role and are not adapted. This default makes unsupported structure
explicit rather than silently broadening the target set.

The parser also constructs a bounded fingerprint
$\boldsymbol{\phi}(G)\in\mathbb{R}^{10}$. Its five core dimensions measure the relative
presence of attention, text fusion, MoE experts, depthwise convolution, and
dense/grouped convolution:
\begin{equation}
\boldsymbol{\phi}_{\rm core}
=(\phi_{\rm attn},\phi_{\rm text},\phi_{\rm moe},\phi_{\rm dw},\phi_{\rm conv}).
\end{equation}
For example, $\phi_{\rm attn}=|\mathrm{attention}|/|V_{\rm role}|$ and
$\phi_{\rm dw}=|\mathrm{depthwise\ conv}|/|V_{\rm role}|$. Five additional statistics
encode depth, width, head-parameter ratio, residual density, and normalization type.
Core dimensions drive stability analysis and variant-level LOVO; the extended
dimensions retain scale diagnostics. Neither is validated on unseen detector families.

\begin{table*}[!t]
\centering
\caption{\textbf{PEFT-on-YOLO measured matrix.} W\&B export on the official
VOC2007 test set (legacy log alias: \texttt{val2007});
anchors are Full-SFT runs of the same backbone. ``rs'' denotes RS-LoRA
scaling; bold values match or exceed Full-SFT. Seed-0 results are shown; selected
core reruns at seeds $\{0,28,42\}$ have sample SD $\leq0.006$ mAP$_{50:95}$
and unchanged rankings. The RT-DETR-L refused row records a within-scope
planner safeguard: all seven swept LoRA-family configurations fall below
$\Delta=-0.05$, so the planner emits \textsc{REFUSE} rather than an adapter;
this is not a held-out-architecture test.
\label{tab:core_wandb}}
\small\renewcommand{\arraystretch}{0.9}\setlength{\tabcolsep}{3pt}
\resizebox{\textwidth}{!}{%
\begin{tabular}{l l c c c c c r r r r r}
\toprule
Backbone & Variant & $r$ & $\alpha$ & rs & DoRA &
\textsc{op-targets} &
mAP$_{50}$ & mAP$_{50:95}$ & $\Delta$ & Total Params & Inference GFLOPs \\
\midrule
\multicolumn{12}{l}{\emph{YOLO11s} (\textsc{dense-conv} only,
$\phi_{\textrm{attn}}{=}0$, anchor $0.6428$)} \\
\midrule
\rowcolor{my_lightgray}
YOLO11s & Full-SFT (anchor)        & --   & --   & --    & --    & all          & $0.8344$ & $0.6428$ & ---       & $9.44$M  & $21.6$ \\
YOLO11s & LoRA                     & $16$ & $32$ & true  & false & conv         & $0.8865$ & $\mathbf{0.7138}$ & $\mathbf{+0.0710}$ & $10.44$M & $25.7$ \\
YOLO11s & DoRA                     & $16$ & $32$ & true  & true  & conv         & $0.8865$ & $\mathbf{0.7138}$ & $\mathbf{+0.0710}$ & $10.44$M & $25.7$ \\
YOLO11s & DoRA (no-rs ablation)    & $16$ & $32$ & false & true  & conv         & $0.8350$ & $0.6479$ & $+0.0050$ & $10.46$M & $25.7$ \\
YOLO11s & LoHa                     & $16$ & $32$ & true  & false & conv         & $0.8620$ & $0.6788$ & $+0.0359$ & $11.45$M & $21.6$ \\
YOLO11s & LoKr                     & $16$ & $32$ & true  & false & conv         & $0.8750$ & $\mathbf{0.7033}$ & $\mathbf{+0.0605}$ & $9.51$M  & $21.6$ \\
YOLO11s & IA$^3$                   & --   & --   & --    & --    & conv         & $0.8717$ & $\mathbf{0.6980}$ & $\mathbf{+0.0552}$ & $9.45$M  & $21.6$ \\
YOLO11s & HRA                      & $16$ & $32$ & true  & false & conv         & $0.9023$ & $\mathbf{0.7276}$ & $\mathbf{+0.0848}$ & $10.24$M & -- \\
\midrule
\multicolumn{12}{l}{\emph{YOLO12s} (\textsc{dense-conv}$+$\textsc{attention},
$\phi_{\textrm{attn}}\!\approx\!0.45$, anchor $0.6662$)} \\
\midrule
\rowcolor{my_lightgray}
YOLO12s & Full-SFT (anchor)        & --   & --   & --    & --    & all          & $0.8532$ & $0.6662$ & ---       & $9.26$M  & $21.6$ \\
YOLO12s & LoRA  ($r{=}8$)          & $8$  & $16$ & true  & false & conv$+$attn  & $0.8974$ & $\mathbf{0.7288}$ & $\mathbf{+0.0626}$ & $9.66$M  & $23.6$ \\
YOLO12s & LoRA  ($r{=}16$)         & $16$ & $32$ & true  & false & conv$+$attn  & $0.9007$ & $\mathbf{0.7307}$ & $\mathbf{+0.0645}$ & $10.06$M & $25.6$ \\
YOLO12s & LoRA  ($r{=}32$)         & $32$ & $64$ & true  & false & conv$+$attn  & $0.9022$ & $\mathbf{0.7363}$ & $\mathbf{+0.0701}$ & $10.85$M & $29.6$ \\
YOLO12s & LoRA  (no-rs ablation)   & $16$ & $32$ & false & false & conv$+$attn  & $0.8853$ & $\mathbf{0.7094}$ & $\mathbf{+0.0432}$ & $10.07$M & $25.6$ \\
YOLO12s & LoRA$+$DoRA, no rs       & $16$ & $32$ & false & true  & conv$+$attn  & \textcolor{red}{$0.8041$} & \textcolor{red}{$0.6112$} & \textcolor{red}{$-0.0550$} & $10.07$M & $25.6$ \\
YOLO12s & LoHa                     & $16$ & $32$ & true  & false & conv$+$attn  & $0.8907$ & $\mathbf{0.7222}$ & $\mathbf{+0.0560}$ & $10.85$M & $21.6$ \\
YOLO12s & IA$^3$                   & --   & --   & --    & --    & conv$+$attn  & $0.8928$ & $\mathbf{0.7210}$ & $\mathbf{+0.0548}$ & $9.27$M  & $21.6$ \\
YOLO12s & HRA                      & $16$ & $32$ & true  & false & conv$+$attn  & $0.9111$ & $\mathbf{0.7453}$ & $\mathbf{+0.0791}$ & $9.91$M  & -- \\
\midrule
\multicolumn{12}{l}{\emph{RT-DETR-l} (pure \textsc{attention},
$\phi_{\textrm{attn}}\!\approx\!0.85$, anchor $0.6833$)} \\
\midrule
\rowcolor{my_lightgray}
RT-DETR-l & Full-SFT (anchor)      & --   & --   & --    & --    & all          & $0.8614$ & $0.6833$ & ---       & $32.85$M & $108.1$\\
RT-DETR-l & LoRA-class refused$^{\ddagger}$ & --   & --   & --    & --    & --           & ---      & ---      & --- (refused) & --- & --- \\
\bottomrule
\end{tabular}}\\[2pt]
\end{table*}

\iffalse
\begin{table}[!t]
\centering
uses a deterministic rank-$16$ Conv-LoRA adapter on YOLO11s and
YOLO12s weights; it validates the merge/export mechanism and is not a
replacement for the trained-checkpoint benchmark. Batch size is one and the
input is $640\times640$. Latency is the synchronized tensor-path median over
10 runs after five warm-up calls; throughput is its reciprocal. The reported
MPS memory is allocator telemetry and is not used for cross-mode ranking.}
\label{tab:mps-deploy}
\scriptsize\setlength{\tabcolsep}{3pt}
\resizebox{\linewidth}{!}{%
\begin{tabular}{l l r r r r r r}
\toprule
Backbone & Mode & Params & Adapter Params & GFLOPs & Latency (ms) & Throughput (img/s) & MPS driver $\Delta$ (MB) \\
\midrule
YOLO11s & unmerged & 10.491M & 1.033M & 25.891 & 27.719 & 36.076 & 1024 \\
YOLO11s & merged & 9.459M & 0 & 21.718 & 17.933 & 55.763 & 1024 \\
YOLO12s & unmerged & 10.380M & 1.094M & 26.199 & 35.095 & 28.494 & 8 \\
YOLO12s & merged & 9.286M & 0 & 21.692 & 24.881 & 40.191 & 8 \\
\bottomrule
\end{tabular}}
\footnotesize The unmerged/merged output maximum absolute error is
$1.16\times10^{-3}$ / $1.22\times10^{-3}$ for YOLO11s/YOLO12s, respectively;
mean absolute error is $1.23\times10^{-6}$ / $1.13\times10^{-6}$. These are
\end{table}
\fi

\subsection{Constraint-Resolved Placement and Safety}
\label{sec:method:planner}
\label{sec:method:guards}

Given the role-aware graph, \textsc{Planner} resolves placement in a fixed order:
\emph{filter operators, filter semantics, apply architecture policies, allocate ranks,
then validate reliability}. Validity precedes efficiency because optimizing a rank
assignment over structurally unsafe layers cannot produce a deployable plan.

\begin{table*}[!t]
\centering
\caption{Formal constraint interface. Each row specifies the information consumed,
the deterministic acceptance rule, the planner output, and the reason returned when
the rule fails. Reliability calibration is evaluated only after the hard constraints.
\label{tab:formal-constraints}}
\scriptsize\setlength{\tabcolsep}{3pt}
\resizebox{\textwidth}{!}{%
\begin{tabular}{p{0.13\textwidth}p{0.20\textwidth}p{0.33\textwidth}p{0.13\textwidth}p{0.17\textwidth}}
\toprule
Constraint & Input & Decision rule & Output if valid & Rejection reason \\
\midrule
Operator validity & $o_i,p,r$: operator type/shape, groups $G$, requested variant and rank
& Dense conv/linear must be supported by the selected backend. Grouped conv requires
group-local factors and $G\mid r$. Depthwise, norm/activation and unknown operators are
excluded unless a structure-preserving implementation is registered.
& Candidate module with a legal adapter shape
& \texttt{E-OP-UNSUPPORTED}, \texttt{E-GROUP-RANK}, \texttt{E-DEPTHWISE-MIX} \\
Semantic safety & $s_i$ and loss-coupled role tags
& Exclude fixed DFL bin projections, MoE routers and geometry-sensitive regression
subpaths. Classification/regression heads are not treated uniformly: only explicitly
whitelisted, loss-compatible subpaths may survive.
& Semantically admissible candidate
& \texttt{E-DFL-FIXED}, \texttt{E-HEAD-GEOMETRY}, \texttt{E-ROUTER-ASSIGN} \\
Graph interface & $q_i$: tensor shapes, fan-in/out, residual and fusion edges
& Substitution must preserve input/output shape, number and ordering of tensors, residual
identity, and the surrounding execution container.
& Interface-preserving host replacement
& \texttt{E-SHAPE}, \texttt{E-MULTIINPUT}, \texttt{E-RESIDUAL} \\
Architecture policy & $\boldsymbol\phi(G),p$ and family-scoped guards
& Apply only policies calibrated for the recognised family. RT-DETR decoder sampling
offsets/attention weights remain frozen; unsupported decoder-wide requests are refused.
YOLO12 attention and YOLO-World fusion use their scoped target policies.
& Restricted candidate set or documented variant rewrite
& \texttt{E-DECODER-CALIB}, \texttt{E-FAMILY-UNKNOWN} \\
Budget feasibility & Candidate costs $c_p(i,r)$ and budget $B$
& Require $\sum_i c_p(i,\pi(i))\le B$ and at least one nonzero-rank target.
& Budget-feasible rank assignment
& \texttt{E-BUDGET}, \texttt{E-EMPTY-PLAN} \\
Deployment compatibility & $h_i$: backend, merge, checkpoint and exporter capabilities
& Installed adapters must support adapter-only persistence and either a verified merge or
an explicitly supported unmerged export path.
& Train--save--merge--export contract
& \texttt{E-NOMERGE}, \texttt{E-EXPORT}, \texttt{E-CHECKPOINT} \\
Reliability calibration & Surviving plan, $\boldsymbol\phi(G)$ and variant $p$
& Within calibrated coverage, refuse if predicted $\Delta$ mAP is below $-0.05$; outside
coverage return unsupported rather than extrapolating a safety guarantee.
& \textsc{Accept} with risk score
& \texttt{R-CATASTROPHE}, \texttt{R-OUT-OF-SCOPE} \\
\bottomrule
\end{tabular}}
\end{table*}

\paragraph{Operator and semantic filtering.}
Depthwise, normalization/activation, and unknown operators are removed first. For a
grouped convolution with $G$ groups, the total rank is distributed as
$r=\sum_{g=1}^{G}r_g$; balanced allocation uses $r_g=r/G$ and therefore requires
$G\mid r$. The planner next removes fixed DFL projections, MoE routers, and
geometry-sensitive regression paths. User targets $T$ and the layer interval $L$ are
intersected only after these mandatory filters, so an explicit request cannot re-enable
an unsafe module.

\paragraph{Architecture-conditioned policies.}
Family-scoped policies encode empirical safeguards, not unseen-family guarantees. The
evaluated RT-DETR-L profile ($\phi_{\rm attn}>0.7$) refuses LoRA-family configurations
below the calibrated threshold; unrecognised Transformer detectors remain unsupported.
On attention-heavy YOLO12, DoRA is converted to LoRA and safe-attention training is
enabled. Rank $8$ is an uncalibrated default, not a cap: the evaluated profile admits
$r\in\{8,16,32\}$ under Eq.~\eqref{eq:budget}, with $r=16$ as the Pareto default.
YOLO-World text-fusion requests can be routed from LoRA to LoHa, while CNN-only graphs
disable attention targets. At module level, YOLO12
excludes \texttt{attn.\{qkv,proj,pe\}} and internal Area-Attention MLP convolutions;
RT-DETR excludes grid-initialised \texttt{sampling\_offsets} and zero-initialised
\texttt{attention\_weights}; and \texttt{*.dfl.*} projections are always frozen. These
rules are deliberately conservative because an unsafe placement is worse than returning
no adapter.

\paragraph{End-to-end example: YOLO12s.}
Consider a request for rank-16 RS-LoRA over all convolutional and linear modules. The
parser first expands the model into backbone, neck and decoupled-head nodes and attaches
operator/semantic roles. The raw name-based request includes ordinary $3\!\times\!3$
convolutions, depthwise convolutions, Area-Attention internals, classification and
regression branches, and the DFL projection. Operator filtering removes depthwise hosts
that lack a channel-preserving adapter (\texttt{E-DEPTHWISE-MIX}); semantic filtering
removes \texttt{*.dfl.*} and geometry-sensitive regression nodes
(\texttt{E-DFL-FIXED}/\texttt{E-HEAD-GEOMETRY}); the YOLO12 policy removes
\texttt{attn.\{qkv,proj,pe\}} and internal Area-Attention MLP convolutions under the
evaluated profile. Graph-interface checks retain shape-preserving backbone/neck hosts.
The budget solver then assigns $r=16$ to the surviving dense convolution targets until
$B$ is met. The emitted target list is the fully qualified names of those surviving
backbone/neck convolutions, accompanied by the excluded-name/reason map and the realised
parameter cost. Thus the final target set is generated by graph roles and constraints,
not by a hidden hand-written list; changing the user range $L$ or budget $B$ changes only
the final intersection and rank assignment, never the mandatory exclusions.

\paragraph{Budgeted rank assignment.}
After filtering, the planner assigns ranks by
\begin{equation}
\begin{aligned}
\max_{\pi}\quad
&\sum_{i\in V_{\rm cand}}u(i,\pi(i);p,\boldsymbol{\phi})\\
\text{s.t.}\quad
&\sum_i c_p(i,\pi(i))\leq B .
\end{aligned}
\label{eq:budget}
\end{equation}
For LoRA-style adapters with $b$ bytes per trainable parameter,
$c_{\rm linear}=br(d_{\rm in}+d_{\rm out})$ and
$c_{\rm conv}=br(C_{\rm in}k_hk_w+C_{\rm out})$. Grouped convolution uses
$b\sum_g r_g((C_{\rm in}/G)k_hk_w+C_{\rm out}/G)$. In rule-only mode,
$u=u_{\rm op}+u_{\rm sem}+u_{\rm range}+u_{\rm rank}-\lambda c_p$; candidate pairs
with non-positive utility are discarded. The resulting plan is refused if every rank is
zero or its realised cost exceeds $B$.

\paragraph{Reliability calibration.}
For candidate pairs not settled by the hard constraints, calibrated mode estimates
\begin{equation}
\Delta\mathrm{mAP}\approx
\beta_0+\beta_1\phi_{\rm attn}+\beta_2\phi_{\rm text}
+\beta_3\phi_{\rm dw}+\beta_4\xi_p ,
\label{eq:regression}
\end{equation}
where $\xi_p$ is a variant coefficient fitted on the canonical matrix. The prediction is
recorded for auditability and triggers refusal when it falls below the catastrophe
threshold. Rule-only mode remains the default; LOVO evaluates calibrated mode on observed
architectures, not unseen families.

% \subsection{YOLO-Compatible Runtime Contract and Metadata}
\subsection{YOLO-Compatible Runtime and Export Contract}
% \label{sec:method:contract}
\label{sec:method:contract}
\label{sec:method:manifest}

A numerically promising placement is not useful if it breaks training or export.
\textsc{Contract} therefore lowers each $(i,r)$ while preserving four invariants:
(I1) base-checkpoint loading is unchanged; (I2) adapter checkpoints store adapter tensors
and runtime metadata only; (I3) merging restores ordinary YOLO modules; and (I4) ONNX
and TensorRT exporters see an export-compatible graph.

The runtime exposes one interface over two backend families. HuggingFace PEFT handles
LoRA, RS-LoRA, DoRA, LoHa, LoKr, AdaLoRA, IA$^3$, OFT, BOFT, and HRA. A narrower
in-repository backend supplies plain convolutional LoRA when PEFT is unavailable or
explicitly bypassed. Both routes share the same resolved target set and lifecycle:
install before optimizer construction, train with frozen base weights, save adapter-only
state, load it onto a compatible base detector, merge, remove wrappers, and then export.
Convolution fusion is disabled while adapters remain unmerged because premature fusion
would invalidate their host modules.

For fallback convolutional LoRA, each group update
$\Delta W_g=(B_gA_g^\top)$ is reshaped to the host kernel. Merging
$W_0\leftarrow W_0+s\Delta W$ therefore preserves the convolution output up to
floating-point operation order.

% \begin{proposition}[Merge equivalence for fallback convolutional LoRA]
% \label{prop:merge}
% For plain LoRA installed through the fallback convolutional backend, merging the
% group-local updates and removing the wrapper preserves the host convolution output up to
% numerical tolerance without mixing channels across groups.
% \end{proposition}

\paragraph{Merge equivalence.}
For fallback convolutional LoRA, each group-local update
$\Delta W_g=B_gA_g^{\top}$ is reshaped to its corresponding host-kernel partition and merged as $W_0\leftarrow W_0+s\Delta W$. Since the updates are constructed independently within each convolution group, removing the adapter wrapper preserves the grouping structure and produces outputs equivalent to those of the wrapped module up to floating-point tolerance. The proof, numerical tolerance checks, and backend-routing policy are included in the supplementary materials. \textsc{Manifest} records backend, variant, targets,
freeze/head settings, and backend-specific files. These fields select the correct loader and detect path confusion; stronger base-checkpoint and class-vocabulary hashes are treated as deployment extensions rather than current guarantees.

\subsection{Training Strategies}
\label{sec:method:training}

Once a plan is accepted, four optional strategies stabilize optimization without changing
placement. \emph{Layer-wise learning-rate decay} uses
$\eta_\ell=\eta_0\rho^{d_\ell}$ with $\rho=0.85$, where normalized depth $d_\ell$
reduces updates in early feature extractors. Similar depth factors are coalesced into a
small number of optimizer groups to avoid per-layer overhead.

\emph{Alpha warmup} ramps LoRA scaling $\alpha/r$ from zero to its target with a cosine
schedule. This prevents a full-strength random adapter from perturbing attention-heavy
detectors at the start of training; the YOLO12 guard requires at least three warmup
epochs. \emph{Orthogonal regularization} adds
\begin{equation}
\lambda_{\rm ortho}
\left(\|A^\top A-I\|_F+\|B^\top B-I\|_F\right)
\end{equation}
every $N$ batches, with $\lambda_{\rm ortho}=0.5$ by default. Its chunked computation
bounds memory when many adapters are active. Finally, \emph{dynamic dropout} linearly
increases the adapter dropout rate from $0$ to $0.15$ after the first $30\%$ of epochs,
preserving early gradient signal while regularizing later updates. Each strategy can be
enabled independently through \texttt{LoRAConfig}; full settings and component
ablations are reported in the supplementary materials.

\section{Experiments}
\label{sec:exp}

\subsection{Implementation Details}
\label{sec:setup}

We follow the official Ultralytics VOC configuration~\citep{everingham2015voc,jocher2023yolov8}:
VOC2007 \texttt{trainval} (5,011) $\cup$ VOC2012 \texttt{trainval} (11,540)
forms the 16,551-image training set, and the official VOC2007 \texttt{test} split contains
4,952 images. The YAML key \texttt{val} resolves to \texttt{images/test2007};
W\&B's historical \texttt{val2007} string is only an alias, not the VOC2007
\texttt{val} directory used in training. Year-prefixed ID intersection is zero
(Supplementary Table~\ref{tab:voc_split_audit}). The study covers CNN detectors (YOLOv8/YOLO11),
attention-augmented YOLO12, text-fusion YOLO-World, and the Transformer-based RT-DETR-L.
Core W\&B-grounded s-scale runs use a $600$-epoch cap and $640\times640$ inputs;
the wider $14$-variant-by-$5$-architecture diagnostic sweep uses $300$ epochs at
$320\times320$. All s-scale ablations use the core protocol; per-run logs retain actual
stopping epochs and batch sizes. For the core runs, Ultralytics'
\texttt{optimizer=auto} resolves to SGD with initial learning rate $0.01$ and
momentum $0.9$; weight decay is $5\times10^{-4}$, cosine scheduling is enabled,
and Mosaic closes for the final ten epochs.
Unless varied, PEFT uses $r=16$, $\alpha=32$, dropout $0.05$, and RS-LoRA scaling.
Experiments use distributed data-parallel training in a consistent
multi-accelerator training environment.
Full-SFT, standard LoRA, Planner-LoRA, best PEFT, and naive placement were evaluated with
seeds $\{0,28,42\}$. Their maximum within-configuration SD was
$\leq0.006$ mAP$_{50:95}$; all directions and rankings were unchanged.
Table~\ref{tab:core_wandb} shows the complete seed-0 matrix; the reruns are a
robustness check, not a paired significance test. All compared core methods share
the evaluation cadence, early-stopping patience (100), and best-checkpoint rule,
so early stopping confers no method-specific advantage. We interpret the scores
as controlled relative comparisons, not once-only blind test estimates.

\begin{table}[!t]
\centering
\caption{\textbf{Structure-aware placement versus LM-inherited defaults.}
The same backbone and evaluation protocol are used in both columns; gains are
the structure-aware minus inherited mAP$_{50:95}$.}
\label{tab:rank_main}
\scriptsize\setlength{\tabcolsep}{3pt}
\resizebox{\columnwidth}{!}{%
\begin{tabular}{l l r l r r}
\toprule
Backbone & LM default & mAP & Planner & mAP & Gain \\
\midrule
YOLO11s & DoRA (no rs) & 0.6479 & RS-LoRA & \textbf{0.7138} & +0.0659 \\
YOLO12s & DoRA (no rs) & 0.6112 & RS-LoRA & \textbf{0.7307} & +0.1195 \\
RT-DETR-L & LoRA sweep & catastrophic & \textbf{REFUSE} & Full-SFT & -- \\
\bottomrule
\end{tabular}
}
\end{table}

\subsection{Analysis of Architecture-Conditioned Behavior}
\noindent\textbf{Macro Stability Mapping.} As illustrated in Figure~\ref{fig:matrix} and Supplementary Table~\ref{tab:matrix}, PEFT reliability is strongly coupled with architecture among the five evaluated detector families rather than obeying a common ranking on this matrix. The catastrophe rate, defined as an mAP drop $\Delta < -0.05$, generally rises with the attention ratio $\phi_{\mathit{attn}}$. Specifically, while standard variants show zero collapse out of ten runs on the evaluated CNN-only YOLO11n configuration, the collapse rate is 6/7 on YOLO12n and 7/7 on RT-DETR-L. The pairwise Kendall-$\tau$ analysis in Supplementary Table~\ref{tab:kendall} likewise shows that rank correlation weakens or reverses across the evaluated families; it is not evidence about every YOLO-family model.

\begin{figure}[!t]
\centering
\includegraphics[width=0.88\linewidth]{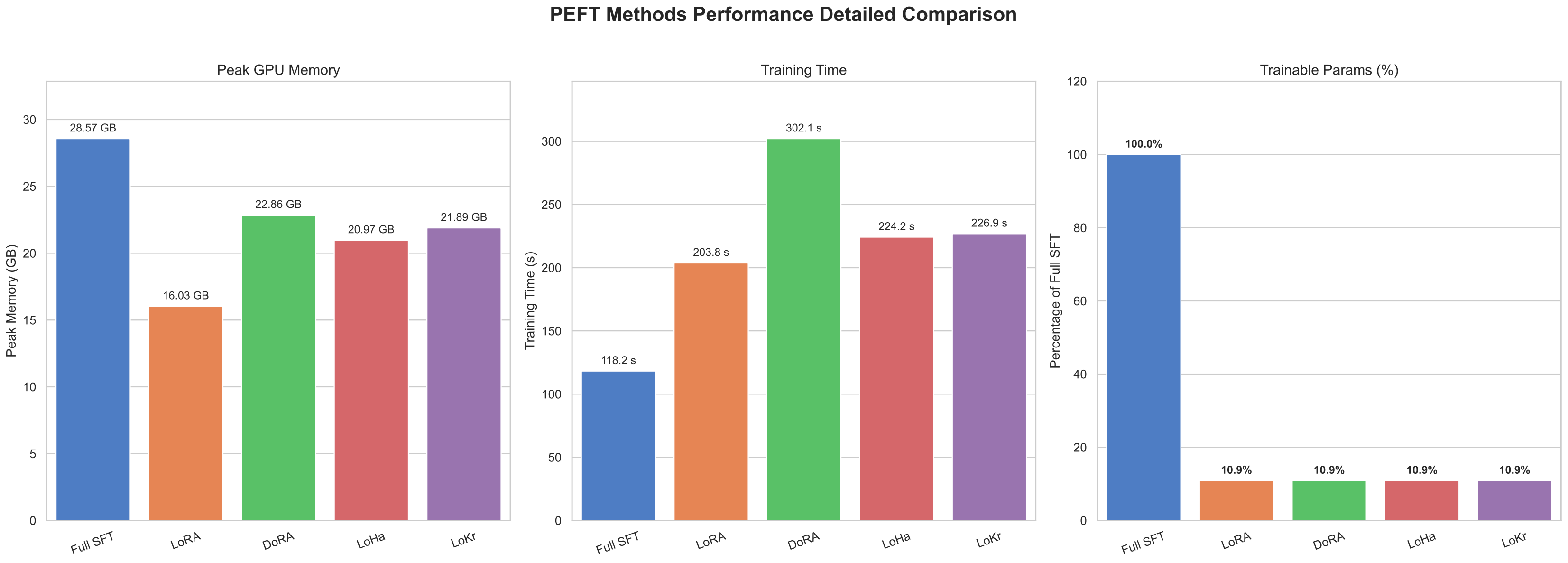}
\caption{\textbf{Controlled PEFT efficiency audit.} On the audited YOLO11
configuration, adapter training reduces peak VRAM but increases wall-clock
time. The figure reports only the controlled measurements used in this paper.}
\label{fig:efficiency}
\end{figure}

\noindent\textbf{Mechanistic Failure Analysis.} We observe that existing PEFT interfaces, which assume homogeneous Transformer blocks, violate the structural constraints of real-time detectors at multiple levels. In multi-modal architectures like YOLO-World, the image and text branches exhibit gradient norms differing by orders of magnitude. Methods like DoRA amplify this mismatch via their magnitude decomposition, leading to rapid loss divergence within 15 epochs and yielding a negative delta of $\Delta = -0.292$. Furthermore, adapting regression heads perturbs fixed-bin DFL projections, which directly conflicts with detection-specific loss constraints and corrupts the loss geometry.

\noindent\textbf{Refusal within Evaluated Coverage.} The RT-DETR-L rule is calibrated on seven observed collapses, not an unseen architecture. LOVO evaluates held-out variants on five known architectures (86.7\% accuracy, $F_1=0.850$).
Treating $\Delta<-0.05$ as unsafe, these seven attempted configurations have a
$7/7$ ($100\%$) unsafe-placement rate. Returning the least-bad adapter would therefore
still be a false acceptance; refusal avoids seven full training attempts once the rule is
calibrated. The current evidence does \emph{not} identify an alternative safe PEFT
configuration for RT-DETR-L. A true false-refusal/false-acceptance rate over unseen
architectures cannot be computed from this matrix because no detector family is held out;
we report this limitation rather than treating variant-level LOVO as architecture-level
validation. Using the controlled run duration as an illustrative opportunity cost, seven
rejected $203.8$-s attempts correspond to approximately $23.8$ minutes of measured
training time; this is a protocol-local estimate, not a general GPU-cost claim.

\subsection{Comparison with Alternative Paradigms}
\noindent\textbf{Comparison with LM-Inherited Rankings.} In Table~\ref{tab:rank_main}, we contrast our structure-aware plan against naive language-model PEFT placement. The displayed YOLO11s/YOLO12s values are the seed-0 core snapshot; across seeds $\{0,28,42\}$, sample SD was $\leq0.006$ mAP$_{50:95}$ and the ranking was preserved. On the evaluated RT-DETR-L configurations, the calibrated planner falls back to Full-SFT rather than returning an adapter below the catastrophe threshold.

\noindent\textbf{Comparison with Full Fine-Tuning (Full-SFT).} Updating all parameters on a mid-scale target dataset can over-adapt pretrained representations, whereas freezing most of the detector and restricting updates to low-rank subspaces can act as an implicit regularizer. On the full VOC trainval split, the strongest YOLO12s configuration (HRA) reaches $0.7453$ $\text{mAP}_{50-95}$ compared with $0.6662$ for Full-SFT, a gain of \textbf{+7.9 points} in the seed-0 core snapshot. Seeds 28 and 42 preserved this direction within the audited SD bound ($\leq0.006$), so this observation is not interpreted as a few-shot result.

\noindent\textbf{MoE stress test.} Table~\ref{tab:moe-stress} reports the supplied
YOLO-Master-EsMoE-S W\&B export. All rows use VOC, $640^2$ input, batch 128, a 600-epoch
cap, the same distributed training environment, seed 0, and disabled planner,
attention, head, router, and expert targeting.
Thus these runs test whether PEFT restricted to ordinary eligible hosts can coexist with
an MoE detector; they do not validate expert-aware planning. HRA is the best completed
configuration at $0.7454$ mAP$_{50:95}$ versus the logged Full-SFT value $0.6891$.
BOFT and OFT fail before producing metrics, while all six metric-bearing PEFT runs finish.

\begin{table*}[!t]
\centering
\caption{Seed-0 YOLO-Master-EsMoE-S stress test from the supplied W\&B export.
Planner and MoE/attention/head targeting are disabled. ``Fail'' denotes a run without
reported evaluation metrics; the logged Full-SFT run reports final metrics although its
W\&B state is \texttt{crashed}. Runtime is wall-clock hours from the export.
\label{tab:moe-stress}}
\small\setlength{\tabcolsep}{4pt}
\begin{tabular}{lrrrrrr}
\toprule
Method & State & Runtime (h) & mAP$_{50}$ & mAP$_{50:95}$ & $\Delta$ & Total Params \\
\midrule
Full-SFT & crashed$^\dagger$ & 5.54 & 0.8719 & 0.6891 & -- & 9.684M \\
LoRA & finished & 2.18 & 0.8687 & 0.6899 & +0.0008 & 10.196M \\
RS-LoRA & finished & 2.14 & 0.8786 & 0.6978 & +0.0087 & 10.196M \\
LoKr & finished & 6.56 & 0.9070 & 0.7402 & +0.0511 & 9.721M \\
LoHa & finished & 2.76 & 0.9062 & 0.7357 & +0.0466 & 10.708M \\
IA$^3$ & finished & 3.88 & 0.9110 & 0.7426 & +0.0535 & 9.693M \\
HRA & finished & 11.37 & \textbf{0.9130} & \textbf{0.7454} & \textbf{+0.0563} & 10.092M \\
BOFT & fail & 0.08 & -- & -- & -- & -- \\
OFT & fail & 0.01 & -- & -- & -- & -- \\
\bottomrule
\end{tabular}
\end{table*}

These data expand the evaluated architecture set, but not the dataset or seed coverage.
They cannot support claims about COCO, domain shift, few-shot learning, or variance. They
also show why a planner needs capability checks: two configurations consume setup time
but never reach evaluation, whereas completed methods differ substantially in runtime.

\noindent\textbf{Disaggregated Deployment Economics.} In our controlled YOLO11
audit, LoRA uses $16.03$ versus $28.57$\,GB ($43.9\%$ less) but takes $203.8$
versus $118.2$\,s ($1.72\times$ longer); other PEFT variants are also slower.
This audit is the primary efficiency claim in this paper; externally
reported storage figures are intentionally excluded. RT-DETR-L's Full-SFT
fallback is excluded from all PEFT aggregates; detailed accounting scope
appears in the supplementary deployment audit.

\noindent\textbf{Interpretation of the trade-off.} The audit separates two
resources that are often conflated. Relative to Full-SFT, the adapter run
removes $12.54$\,GB of peak training memory but adds $85.6$\,s of training
time on the same controlled configuration. The memory reduction therefore
comes with an optimization-time cost, rather than constituting a universal
speedup. Deployment has a separate choice: the unmerged adapter is useful for
checkpoint interchange, whereas the merged model removes the adapter branch
before export and recovers the base-model operator path (Supplementary
Table~\ref{tab:mps-deploy}). We consequently report VRAM, wall-clock time,
and merged/unmerged behavior as separate axes, and treat \textsc{Refuse}$\to$
Full-SFT as a valid outcome when the adapter risk is too high.

% \begin{figure}[!t]
% \centering
% \includegraphics[width=0.88\linewidth]{Figs/ultimate_benchmark.png}
% \caption{\textbf{Controlled PEFT efficiency audit.} On the audited YOLO11
% configuration, adapter training reduces peak VRAM but increases wall-clock
% time. The figure reports only the controlled measurements used in this paper.}
% \label{fig:efficiency}
% \end{figure}

% \begin{figure}[!t]
% \centering
% \includegraphics[width=0.88\linewidth]{Figs/ultimate_benchmark.png}
% \caption{\textbf{Controlled PEFT efficiency audit.} On the audited YOLO11
% configuration, adapter training reduces peak VRAM but increases wall-clock
% time. The figure reports only the controlled measurements used in this paper.}
% \label{fig:efficiency}
% \end{figure}

\begin{figure}[!t]
\centering
\includegraphics[width=0.92\linewidth]{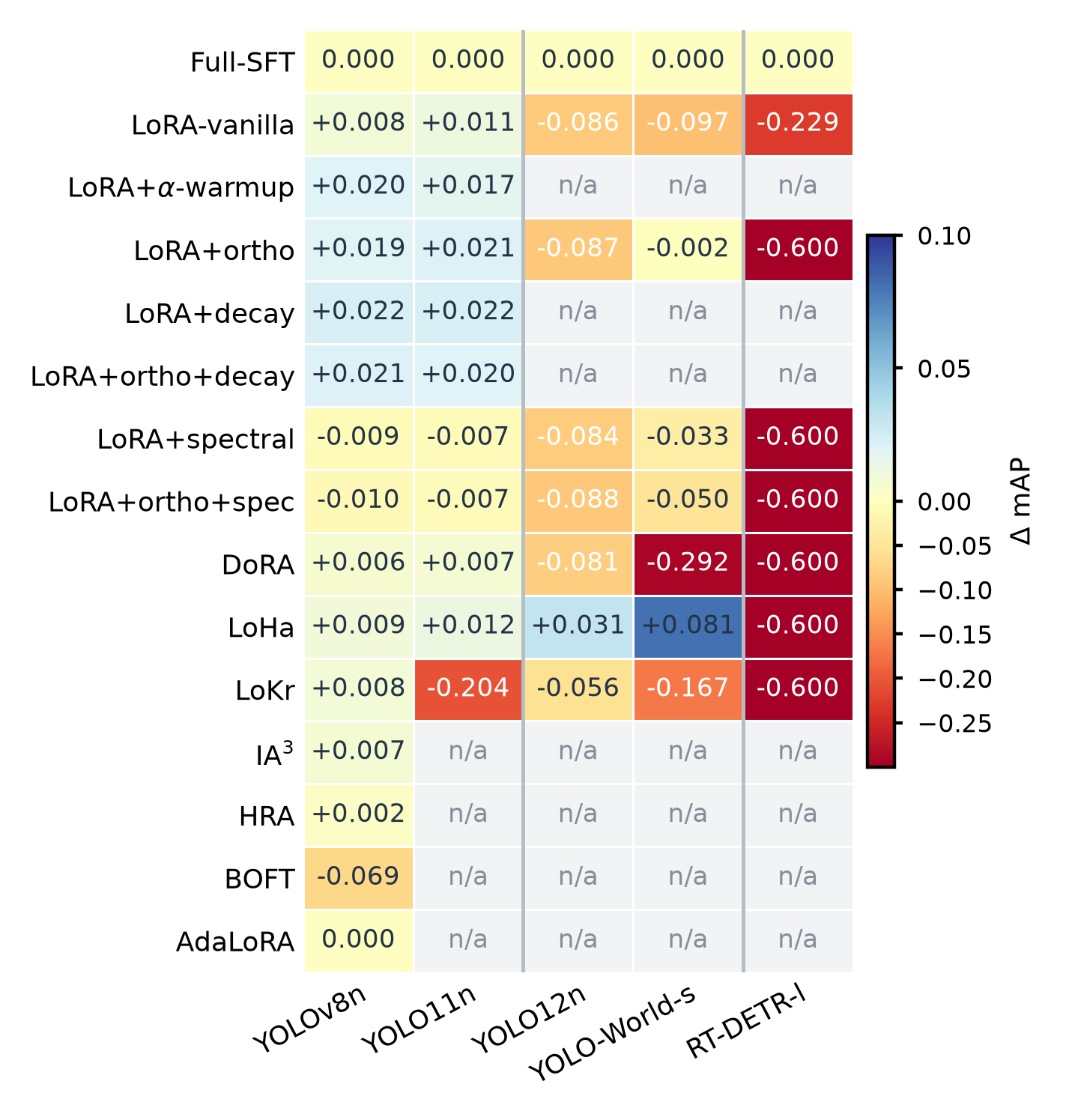}
\caption{\textbf{Architecture-conditioned PEFT behavior.} $\Delta$mAP
versus Full-SFT for 14 variants on five heterogeneous detectors. Red cells
are catastrophic ($\Delta<-0.05$); gray denotes not swept and colors clip below
$-0.30$. The catastrophe rate rises with $\phi_{\mathrm{attn}}$, and no common ranking survives across the five evaluated families.
\label{fig:matrix}}
\end{figure}

\subsection{Ablation Study}
\noindent\textbf{Effectiveness of Planner Constraints.} We isolate the performance contributions of individual planner components on YOLO12s. As detailed in Supplementary Table~\ref{tab:ablation_router}, a naive baseline without constraints yields $0.6900$ $\text{mAP}_{50-95}$. Introducing operator validity filtering to exclude depthwise and normalization layers elevates the accuracy to $0.7094$. Incorporating detection-head semantic exclusions provides the largest stability leap, driving the performance to \textbf{0.7307 mAP}, which supports semantic guards as stability drivers rather than mere engineering safeguards.

\begin{table*}[!t]
\centering
\caption{Constraint diagnostics on the evaluated YOLO12s and RT-DETR-L graphs.
``Added'' counts modules admitted when one predicate is removed. Gradient and
$\Delta$ mAP values are seed-0, one-epoch MPS diagnostics on 2\% of VOC07+12
with a fixed 160-image validation subset; they are not main-protocol performance
results. DFL and RT-DETR risky layers are counterfactual audits only.
\label{tab:constraint-diagnostics}}
\small\setlength{\tabcolsep}{4pt}
\resizebox{\textwidth}{!}{%
\begin{tabular}{llrrrrl}
\toprule
Graph & Removed predicate & Safe & Added & Grad. ratio & $\Delta$ mAP$_{50:95}$ & Outcome / reason \\
\midrule
YOLO12s & Attention stream & 55 & 32 & $1.581\times$ & $-0.0113$ & finite; 16 attention + 16 block-MLP \\
YOLO12s & Typed detection head & 55 & 18 & $1.091\times$ & $-0.0337$ & finite; head grad. norm $15.94$ \\
YOLO12s & Depthwise validity & 55 & 0 & -- & -- & not triggered at rank 4 \\
YOLO12s & DFL semantics & 55 & 1 & -- & -- & fixed integral projection; not trained \\
RT-DETR-L & MSDeform geometry & 95 & 12 & -- & -- & sampling grid / softmax; not trained \\
RT-DETR-L & Score/bbox outputs & 95 & 28 & -- & -- & output-bias semantics; not trained \\
\bottomrule
\end{tabular}}
\end{table*}

\noindent\textbf{Predicate-level diagnostics.}
Table~\ref{tab:constraint-diagnostics} separates an applicable constraint from a
constraint that is merely present in the rule set. On YOLO12s, opening the
attention-stream rule admits 32 modules and raises the mean adapter-gradient norm
by $1.581\times$; opening the typed-head rule admits 18 modules and raises it by
$1.091\times$. Neither short run produces non-finite gradients, so the evidence
supports changed optimization pressure rather than inevitable collapse.
Conversely, the depthwise predicate admits no additional rank-4 target on either
audited graph and therefore cannot be credited with a training effect here. The
single DFL projection, 12 deformable-sampling/attention-weight layers, and 28
RT-DETR score/bbox output layers remain counterfactual candidates: their
fixed-bin, sampling-geometry, or output-bias semantics make direct insertion a
different intervention, not an ordinary leave-one-out switch.

\noindent\textbf{Downstream decoder drift under safe placement.}
We compare an RT-DETR-L VOC checkpoint with its safe adapter active against the
same checkpoint with adapter parameters zeroed. Through the supported predictor
path, 16 fixed VOC2007 images produce finite $300\!\times\!6$ outputs. Relative
$\ell_2$ drift is $0.553\!\pm\!0.099$ in decoder modules,
$0.530\!\pm\!0.103$ in MSDeformAttn, $0.613\!\pm\!0.100$ in bbox heads, and
$0.127\!\pm\!0.033$ in score heads; final-output drift is
$0.548\!\pm\!0.094$ (median $0.571$, P90 $0.611$). Safe upstream adaptation
therefore propagates into the frozen decoder, but these 16-image, 1\%-training
diagnostics do not validate unsafe decoder insertion or replace the core protocol.

\noindent\textbf{Rank Sensitivity Analysis.} Supplementary Table~\ref{tab:rank_sweep} evaluates the accuracy-cost trade-offs across different rank budgets where $r \in \{8, 16, 32\}$ on YOLO12s. Increasing the rank from 8 to 16 improves the mAP from $0.7288$ to $0.7307$. Scaling further to $r=32$ offers diminishing returns yielding $0.7363$ mAP at a substantially higher parameter and computational overhead, supporting $r=16$ as a practical Pareto point for real-time detectors.

\noindent\textbf{Necessity of Runtime Contract.} We ablate the contract layer against a naive wrapper substitution. While both approaches reach an identical $0.7138$ mAP with YOLO11s, Supplementary Table~\ref{tab:ablation_hook} demonstrates that the naive substitution breaks downstream deployment pipelines, including ONNX/TensorRT export and standard checkpoint saving. Our \texttt{Contract} layer satisfies the checkpoint, merge, and ONNX invariants; after merging, the adapter branch is removed, so the exported graph has base-model operator cost. 
% The unmerged graph necessarily incurs adapter-branch overhead, quantified in the deployment audit (Supplementary Table~\ref{tab:mps-deploy}).

\noindent\textbf{Interaction of Training Priors.} We conduct a $2 \times 2$ factorial ablation on YOLO12s to examine the interplay between RS-LoRA scaling and DoRA's magnitude decomposition. Supplementary Table~\ref{tab:toxic} reveals that training priors interact non-additively, showing that enabling DoRA without RS-LoRA scaling triggers a collapse to $0.6112$ mAP. This empirical finding underscores the need to report the joint training configuration rather than isolated prior settings. Results are therefore interpreted by protocol: the core matrix, diagnostic architecture sweep, and controlled VRAM/time audit are not pooled, and the planner treats disabled-expert MoE tests and RT-DETR-L refusal as separately scoped outcomes.

\ifdefined\SUPPLEMENTONLY
\begin{figure}[!t]
\centering
\includegraphics[width=0.96\linewidth]{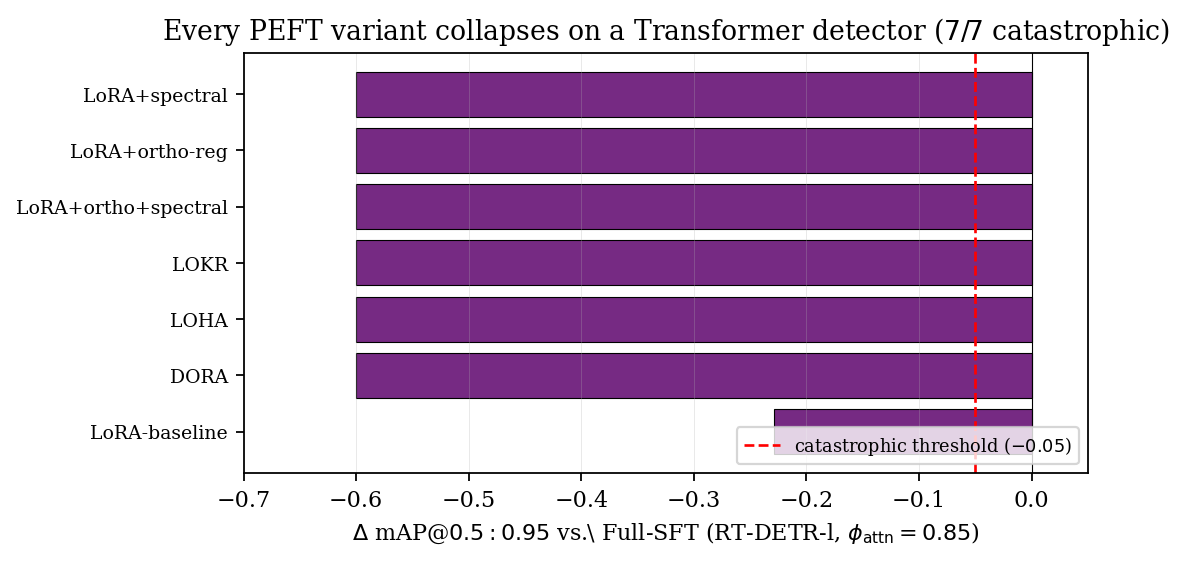}
\caption{\textbf{Refusal prevents catastrophic detector adaptation.}
On RT-DETR-l, every PEFT variant that survives training falls
below the catastrophic threshold; the planner refuses and falls
back to Full-SFT.
\label{fig:rtdetr_collapse}}
\end{figure}
\fi

% In this regime, returning even the least-bad swept adapter would still produce an unusable detector. \system{}'s planner therefore fires its refusal mechanism and falls back to Full-SFT rather than silently shipping a degraded model --- refusal here is not a limitation of the method, but a deliberate and necessary design decision.

\ifdefined\SUPPLEMENTONLY
\ifdefined\SUPPLEMENTONLY
\begin{table}[!t]
\centering
\caption{Architecture-family coverage on PASCAL VOC. CNN rows
aggregate YOLOv3/v5/v6/v8/v9/v10/YOLO11; the full eleven-backbone
coverage table is retained in the supplementary material.
Rows above the final divider report planner outcomes.
$^{\ddagger}$RT-DETR-l: the planner \emph{refuses} all $14$ variants.
$^{*}$MoE detector is a separate planner-disabled stress test.
\label{tab:eleven}}
\small\setlength{\tabcolsep}{3pt}
\resizebox{\columnwidth}{!}{%
\begin{tabular}{lllll}
\toprule
Architecture class & Backbones & Decision & Adapter & Status \\
\midrule
CNN & YOLOv3/v5/v6/v8/v9/v10/YOLO11 & LoRA$+$decay & 1.5--4.0\,MB & accept, $+0.018$--$+0.022$ \\
CNN$+$attention & YOLO12 & LoHa & 2.5\,MB & accept, $+0.031$ \\
Text-fusion & YOLO-World & LoHa & 2.1\,MB & accept, $+0.081$ \\
Transformer detector$^{\ddagger}$ & RT-DETR & \textbf{REFUSE} & -- & avoid $-0.229$ \\
\midrule
MoE$^{*}$ & YOLO-Master & exploratory & -- & expert-aware planning deferred \\
\bottomrule
\end{tabular}
}
\end{table}
\fi
\fi

\ifdefined\SUPPLEMENTONLY
\begin{table}[!t]
\centering
\caption{Predictive validation summary for
Eq.~\eqref{eq:regression}. The main quantity is leave-one-variant-out
catastrophe prediction on unseen variants
($\Delta\!<\!-0.05$).
\label{tab:scaling}}
\small\setlength{\tabcolsep}{4pt}
\begin{tabular}{lr}
\toprule
Metric & Value \\
\midrule
$R^2$ on fitted matrix              & $0.762$ \\
\textsc{lovo} catastrophe accuracy  & $86.7\%$ \\
\textsc{lovo} catastrophe recall    & $0.944$ \\
\textsc{lovo} catastrophe $F_1$     & $0.850$ \\
\bottomrule
\end{tabular}
\end{table}
\begin{figure}[!t]
\centering
\includegraphics[width=0.85\linewidth]{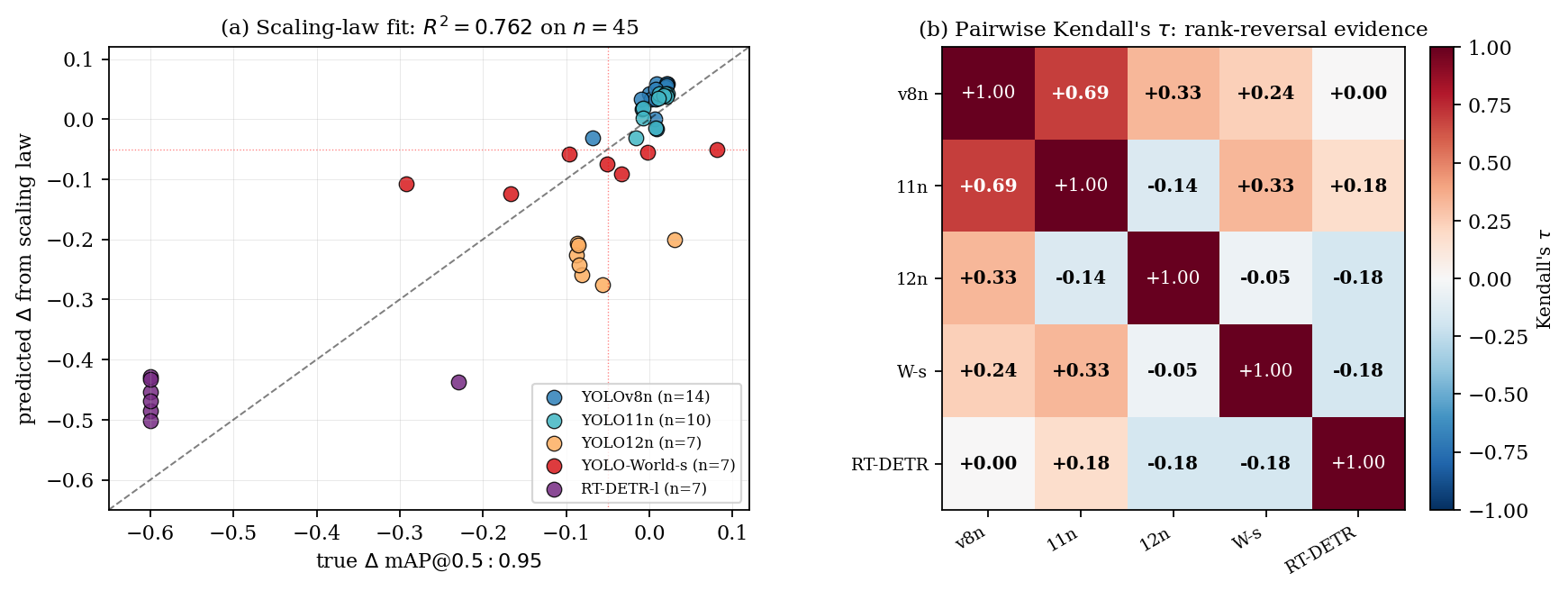}
\caption{\textbf{Vision PEFT Scaling Law fit and pairwise
$\tau$.} \textbf{(a)} Predicted vs.\ true $\Delta$ mAP for all
$45$ cells; the dashed diagonal is the identity, the red dotted
lines mark the catastrophe threshold $\Delta\!=\!-0.05$.
\textbf{(b)} Pairwise Kendall's $\tau$ heatmap.
\label{fig:scaling_law}}
\end{figure}
\fi

\ifdefined\SUPPLEMENTONLY
\begin{table}[!t]
\centering
\caption{Compact ablation suite. Each row compresses one detailed
appendix ablation into the design question it answers.
\label{tab:compact_ablation}}
\scriptsize\setlength{\tabcolsep}{3pt}
\begin{tabularx}{\linewidth}{@{}>{\raggedright\arraybackslash}p{0.27\linewidth}>{\raggedright\arraybackslash}X>{\raggedright\arraybackslash}p{0.25\linewidth}@{}}
\toprule
Design question & Key result & Conclusion \\
\midrule
Contract necessity
& Naive substitution and \modC{} reach the same mAP ($0.7138$); only \modC{} passes ONNX/TRT/Ckpt
& Runtime contract is required \\
Planner constraints
& $0.6900 \rightarrow 0.7094 \rightarrow 0.7307$
& Semantic filtering is the main stability driver (Finding 5) \\
Rank choice
& $0.7288 / 0.7307 / 0.7363$
& $r{=}16$ is a Pareto default, not the accuracy optimum (Finding 6) \\
Variant dependence
& HRA/LoRA/LoHa ordering shifts
& No universal PEFT ranking (Finding 2) \\
RS-LoRA $\times$ DoRA
& DoRA without RS collapses to $0.6112$
& Training priors interact non-additively \\
\bottomrule
\end{tabularx}
\end{table}
\fi

\ifdefined\SUPPLEMENTONLY
\begin{table}[!t]
\centering
\caption{Ablation A7 (key result): LM-inherited ranking vs.\
structure-aware plan. Numbers from W\&B export, official VOC2007 test,
$r{=}16$, $\alpha{=}32$, core $600$-epoch-cap/$640$-resolution protocol.
Gain is the $\Delta$ between
the structure-aware plan and the LM-inherited plan on the same
backbone.
\label{tab:ablation_archcond}}
\small\setlength{\tabcolsep}{3pt}
\resizebox{\columnwidth}{!}{%
\begin{tabular}{l l r l r r}
\toprule
\multirow{2}{*}{Backbone} & \multicolumn{2}{c}{LM-inherited plan}
 & \multicolumn{2}{c}{Structure-aware plan} & \multirow{2}{*}{Gain} \\
\cmidrule(lr){2-3}\cmidrule(lr){4-5}
& Variant & mAP$_{50:95}$ & Variant & mAP$_{50:95}$ & \\
\midrule
YOLO11s   & DoRA (no rs)   & $0.6479$
          & LoRA (rs, planner)  & $\mathbf{0.7138}$ & $+0.0659$ \\
YOLO12s   & DoRA (no rs)   & \textcolor{red}{$0.6112$}
          & LoRA (rs, planner)  & $\mathbf{0.7307}$ & $+0.1195$ \\
RT-DETR-l & LoRA-class sweep & $7/7$ catastrophic
          & \textbf{REFUSE} (Full-SFT, $0.6833$) & --- & --- \\
\bottomrule
\end{tabular}
}
\end{table}
\fi

\ifdefined\SUPPLEMENTONLY\else\vspace{-4pt}\fi
\FloatBarrier
\section{Conclusion}
\label{sec:conclusion}

% TODO: Write conclusion
% - Summarize contributions:
%   1. New formulation of YOLO-family PEFT as structure-constrained adapter placement
%   2. Structure-aware planning and runtime framework
%   3. Empirical finding that adapter stability is architecture-conditioned
% - Discuss limitations and future work
% - Reiterate practical impact

% This paper proposed YOLO-PEFT, a structure-aware PEFT framework that resolves
% YOLO-family PEFT as constrained adapter placement on heterogeneous detector graphs.
% Our framework formulates adapter placement as a constrained planning problem,
% introduces a structure-aware planner with refusal capability, and provides a
% YOLO-compatible runtime for train--save--merge--export workflows.
% Experiments demonstrate significant parameter compression, memory reduction,
% and cost savings while maintaining accuracy across diverse YOLO architectures.

% \section{Conclusion and Future Work}
% \label{sec:conclusion}

We presented \system{}, which parses detector graphs from the supported, evaluated
families, resolves
adapter placements under operator, semantic, and budget constraints, and
lowers accepted plans into a mergeable and export-compatible runtime. Under the
official VOC07+12 trainval$\rightarrow$VOC07 test protocol, accepted YOLO11s
and YOLO12s plans match or exceed Full-SFT. Our controlled YOLO11 audit measures
$43.9\%$ lower LoRA peak VRAM but $1.72\times$ longer training. Externally
sourced storage and distribution estimates are intentionally excluded from
the core claims. RT-DETR-L's \textsc{Refuse}$\rightarrow$Full-SFT fallback is
excluded from these PEFT aggregates. Stability remains architecture-conditioned,
and held-out-architecture validation of refusal remains future work.

The supplied seed-0 YOLO-Master-EsMoE-S export adds a scoped MoE stress test: with
planner, router, expert, attention, and head targeting disabled, HRA reaches $0.7454$
mAP$_{50:95}$ versus the logged Full-SFT value $0.6891$, whereas BOFT and OFT terminate
without metrics. This supports compatibility with a conv-only MoE host subset, not an
expert-aware planner claim. More broadly, the current evidence is limited to VOC and the
evaluated families; COCO/domain-shift evaluation, region-only fine-tuning baselines, and
held-out-family false-refusal/false-acceptance estimates remain necessary.

For practice, the result is a decision rule within the implemented constraints and
calibration coverage, rather than a universal adapter recommendation. A planner-selected
adapter should be used only after operator and semantic checks pass; refusal before
training is supported only for evaluated or calibrated architecture--method regimes.
Future work will extend this rule to VLM detectors with cross-modal
attention and vocabulary heads, and to mixture-of-transformers (MoT/MoE)
detectors with shared blocks, expert-local adapters, router load balancing, and
dispatch-aware budgets. We will test these extensions on held-out architectures
and cross-domain benchmarks, including merged and unmerged export checks.

\FloatBarrier
\clearpage
\appendix
\section{Deployment and Export Audit}
\label{app:deployment-audit}

The following figure and smoke-test table document the controlled deployment
audit and merged/unmerged export checks omitted from the seven-page main paper.

\begin{figure}[!t]
\centering
\includegraphics[width=0.80\linewidth]{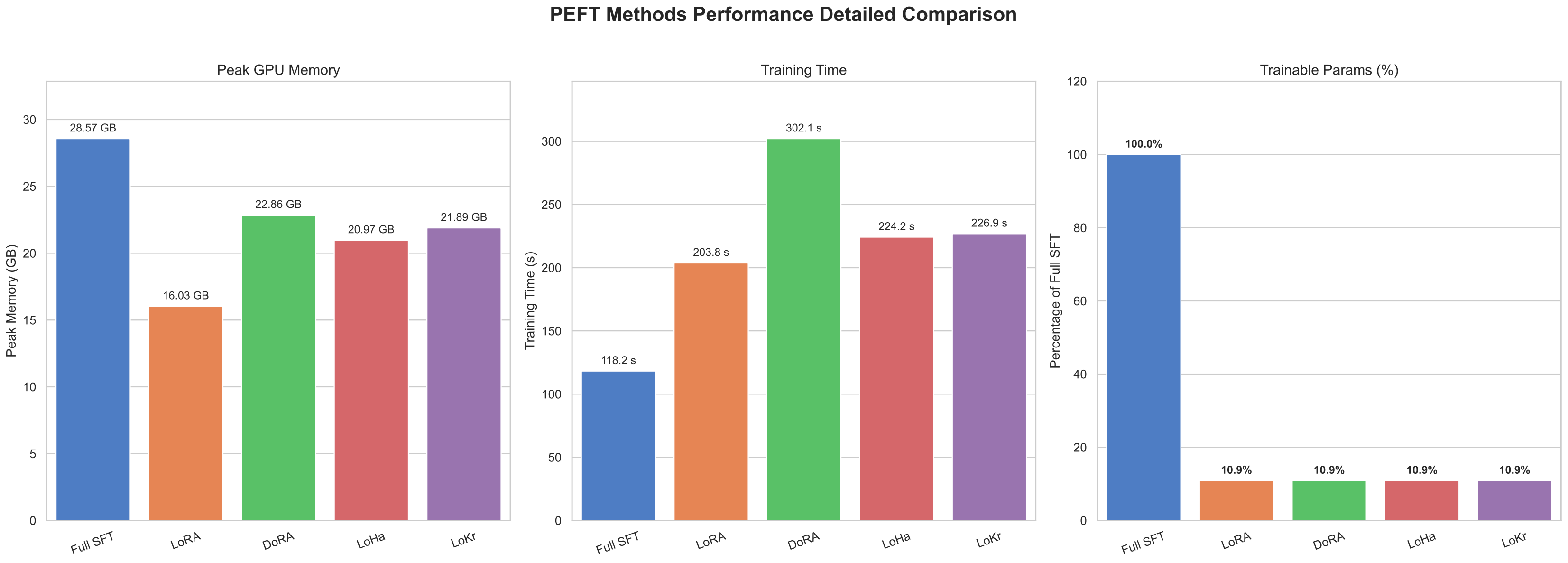}
\caption{Paper-measured controlled YOLO11 systems audit. Relative to Full-SFT,
LoRA lowers peak VRAM from $28.57$ to $16.03$\,GB ($43.9\%$) but increases
training time from $118.2$ to $203.8$\,s ($1.72\times$); all displayed PEFT
variants are slower than Full-SFT.\label{fig:benchmark}}
\end{figure}

\begin{table}[!t]
\centering
\caption{MPS deployment audit on controlled YOLO11s and YOLO12s checkpoints.}
\label{tab:mps-deploy}
\scriptsize\setlength{\tabcolsep}{3pt}
\resizebox{\linewidth}{!}{%
\begin{tabular}{l l r r r r r r}
\toprule
Backbone & Mode & Params & Adapter Params & GFLOPs & Latency (ms) & Throughput (img/s) & MPS $\Delta$ (MB) \\
\midrule
YOLO11s & unmerged & 10.491M & 1.033M & 25.891 & 27.719 & 36.076 & 1024 \\
YOLO11s & merged & 9.459M & 0 & 21.718 & 17.933 & 55.763 & 1024 \\
YOLO12s & unmerged & 10.380M & 1.094M & 26.199 & 35.095 & 28.494 & 8 \\
YOLO12s & merged & 9.286M & 0 & 21.692 & 24.881 & 40.191 & 8 \\
\bottomrule
\end{tabular}}
\end{table}

\paragraph{ONNX/TensorRT verification protocol.}
Both unmerged and merged graphs passed ONNX checker validation at opset 18
for static shape $[1,3,640,640]$ and dynamic batch, height, and width axes;
ONNX Runtime CPU execution gave maximum absolute errors below
$1.24\times10^{-3}$ relative to PyTorch. TensorRT engines were built and
executed in FP32 and FP16 with dynamic optimization profiles, verifying
numerical agreement, memory, throughput, and latency.

% ---- Inlined source: Sec/6_appendix.tex ----
% ----------------------------------------------------------------------------
\section{Merge Equivalence of the Fallback Manual Conv2d Backend}
\label{app:merge-equivalence}
% -----------------------------------------------------------------------------

\subsection{Dense convolution as im2col matrix multiplication}

The fallback manual backend keeps the host \texttt{Conv2d} frozen and
computes an additive LoRA branch on the unfolded input patches.
For a dense convolution, \texttt{unfold} maps
$x\in\mathbb{R}^{N\times C_{\textrm{in}}\times H\times W}$ to
$\tilde{x}\in\mathbb{R}^{N\times C_{\textrm{in}}k_hk_w\times L}$,
where $N$ is the batch size and $L=H'W'$.
The unfold operator uses the same kernel size, stride, padding, and dilation
as the host convolution.
Following the im2col convention, the convolution over unfolded patches is
computed as $\tilde{x}^{\top}W^{\top}$, producing an
$L\times C_{\textrm{out}}$ output before reshaping.
A LoRA branch with rank $r$ computes $(\tilde{x}^{\top}A)B^{\top}$ and
reshapes the result back to the convolutional output layout.
If the host convolution carries a bias term, the bias is preserved unchanged
during merging.

\subsection{Grouped convolution and per-group rank allocation}

For a grouped convolution with $G$ groups, the fallback manual backend uses
per-group factors
\[
  A_g\in\mathbb{R}^{(C_{\textrm{in}}/G)k_hk_w\times r_g},\qquad
  B_g\in\mathbb{R}^{C_{\textrm{out}}/G\times r_g},
\]
with total rank budget $r=\sum_g r_g$.
The implementation uses the balanced allocation $r_g=r/G$, so $r$ must be
divisible by $G$.
This avoids mixing channels across groups: because each group operates on a
disjoint input-channel block, concatenating per-group updates yields a
block-diagonal update in the dense im2col representation, thereby preserving
the host convolution's group structure.
The dense case is recovered by setting $G=1$ and $r_1=r$.

\subsection{Proof of Proposition 1}

For each group $g$, the LoRA branch is linear in the unfolded patches with
effective matrix $B_gA_g^\top$.
Reshaping this matrix to
$(C_{\textrm{out}}/G,\,C_{\textrm{in}}/G,\,k_h,\,k_w)$ yields
\begin{equation}
  \Delta W_g = \bigl(B_gA_g^\top\bigr)
  \ \text{reshaped to}\ (C_{\textrm{out}}/G,\,C_{\textrm{in}}/G,\,k_h,\,k_w).
  \label{eq:merge}
\end{equation}
Concatenating the per-group updates gives a convolutional weight increment
$\Delta W$ of the same shape as $W_0$.
Therefore
\[
  \texttt{conv}(W_0,x)+s\,g((A_g,B_g)_g,x)
  =\texttt{conv}(W_0+s\Delta W,x),
\]
up to floating-point operation ordering.
After applying $W_0\leftarrow W_0+s\Delta W$, the wrapper can be replaced
by a plain \texttt{Conv2d}.

\subsection{Numerical tolerance and implementation notes}

The proof above covers the plain fallback \texttt{Conv2d} LoRA path.
RS-LoRA and PEFT-managed variants are delegated to the PEFT runtime;
few-shot and adaptive fallback variants require the rank mask to be resolved
before applying the same merge pattern.
The merged and unmerged paths differ only in operation ordering
(\texttt{unfold} + batched matrix multiplication versus a single convolution
on the merged weight); outputs are treated as equivalent within standard
floating-point tolerance.
Merge correctness is verified after wrapper removal, not before.
The equivalence holds in deployment/evaluation mode, where adapter dropout
is disabled.

% -----------------------------------------------------------------------------
\section{Backend Routing and PEFT-Managed Variants}
\label{app:backend-routing}
% -----------------------------------------------------------------------------

\subsection{Backend selection rules}

The PEFT backend is the default route for LoRA, RS-LoRA, DoRA, LoHa, LoKr,
AdaLoRA, IA$^3$, OFT, BOFT, and HRA.
The fallback manual backend is intentionally narrower: it is used for plain
\texttt{Conv2d} LoRA wrapping when PEFT is unavailable, explicitly bypassed,
or a fallback path is requested.
Quantized paths are delegated to the PEFT backend because quantization
integration is backend-specific.

\subsection{Quantized and wrapper-required paths}

Wrapper-required training states are not considered export-compatible
artifacts.
If a wrapper cannot be merged into a plain YOLO module or handled by a
PEFT-managed export path, the contract layer refuses export.

\subsection{Export policy}

Export is allowed only when the resulting model satisfies the runtime
invariants stated in the method: unchanged base-checkpoint loading,
adapter-only checkpoints, ordinary YOLO module structure after merge, and
export-compatible modules for ONNX / TensorRT tooling.
A PEFT-managed adapter is export-compatible only after the PEFT merge path
succeeds or after the contract verifies an equivalent unwrapped module
structure.

% -----------------------------------------------------------------------------
\section{Runtime Metadata Schema}
\label{app:manifest-schema}
% -----------------------------------------------------------------------------

\subsection{JSON fields}

The current implementation persists runtime metadata rather than a full
cryptographic manifest.
The PEFT path writes \texttt{runtime\_metadata.json} alongside the PEFT
\texttt{save\_pretrained} artifact; the fallback path writes
\texttt{fallback\_meta.json} and a fallback weight file.
Recorded fields include backend, variant, freeze-BN setting,
head-inclusion setting, target modules, and backend-specific runtime
metadata.

\subsection{Compatibility checks}

At load time, \system{} uses this metadata to select the PEFT or fallback
loader and to verify that the requested backend and target configuration are
consistent with the saved adapter.
Full base-checkpoint, model-YAML, class-name, and architecture-hash checks
are deploy-time extensions rather than current hard-checking requirements.

\subsection{Failure modes}

The metadata check prevents backend/path confusion and missing fallback
weights.
Stronger graph, class-vocabulary, and export-runtime mismatch checks require
the full deploy-time manifest extension described in the preceding subsection.

% -----------------------------------------------------------------------------
\section{PEFT Matrix Details}
\label{app:full-matrix}
% -----------------------------------------------------------------------------

The full core W\&B measured matrix is reported as Table 1
in the main text.
To avoid duplicating that table, this appendix provides the numerical
companion to Fig. 2 and the pairwise rank-agreement
diagnostic.

\subsection{Numerical companion to Fig. 2}

Table S2 reports the compact $\Delta$mAP view corresponding
to the heatmap.
It separates the full-SFT anchor from the PEFT deltas and marks
catastrophic cells.

\begin{table}[!t]
\centering
\caption{Numerical companion to Fig. 2.
  Results are from the $320$-resolution diagnostic matrix.
  The Full-SFT row reports the anchor mAP@$0.5{:}0.95$; all other rows
  report $\Delta$.
  Red entries are catastrophic ($\Delta\!<\!-0.05$).
  For collapsed RT-DETR-l runs, $\Delta=-0.600$ denotes the clipped
  failure value used for visualization. This entire matrix is a diagnostic
  seed-0 sweep; it is not a multi-seed aggregate. The seeds-28/42 reruns
  reported in the main text were used as robustness checks for the core
  baselines and selected planner configurations only.
\label{tab:matrix}}
\scriptsize\setlength{\tabcolsep}{3pt}
\resizebox{\columnwidth}{!}{%
\begin{tabular}{lrrrrr}
\toprule
Variant & YOLOv8n & YOLO11n & YOLO12n & YOLO-W-s & RT-DETR-l \\
\midrule
Full-SFT (anchor)    & $0.586$ & $0.588$ & $0.558$ & $0.491$ & $0.601$ \\
LoRA-vanilla         & $+0.008$ & $+0.011$ & \textcolor{red}{$-0.086$} & \textcolor{red}{$-0.097$} & \textcolor{red}{$-0.229$} \\
LoRA+$\alpha$-warmup & $+0.020$ & $+0.017$ & -- & -- & -- \\
LoRA+ortho           & $+0.019$ & $+0.021$ & \textcolor{red}{$-0.087$} & $-0.002$ & \textcolor{red}{$-0.600$} \\
LoRA+decay           & $+0.022$ & $+0.022$ & -- & -- & -- \\
LoRA+ortho+decay     & $+0.021$ & $+0.020$ & -- & -- & -- \\
LoRA+spectral        & $-0.009$ & $-0.007$ & \textcolor{red}{$-0.084$} & $-0.033$ & \textcolor{red}{$-0.600$} \\
DoRA                 & $+0.006$ & $+0.007$ & \textcolor{red}{$-0.081$} & \textcolor{red}{$-0.292$} & \textcolor{red}{$-0.600$} \\
LoHa                 & $+0.009$ & $+0.012$ & $+0.031$ & $+0.081$ & \textcolor{red}{$-0.600$} \\
LoKr                 & $+0.008$ & \textcolor{red}{$-0.204$} & \textcolor{red}{$-0.056$} & \textcolor{red}{$-0.167$} & \textcolor{red}{$-0.600$} \\
IA$^3$               & $+0.007$ & -- & -- & -- & -- \\
HRA                  & $+0.002$ & -- & -- & -- & -- \\
BOFT                 & \textcolor{red}{$-0.069$} & -- & -- & -- & -- \\
AdaLoRA              & $0.000$ & -- & -- & -- & -- \\
\midrule
Catastrophe rate     & $1/14$ & $0/10$ & $6/7$ & $4/7$ & $7/7$ \\
\bottomrule
\end{tabular}
}
\end{table}

\subsection{Pairwise rank agreement}

Table S3 quantifies the rank-reversal pattern visible in
the supplementary scaling-law figure.
The within-CNN comparison is the only strongly positive pair; crossing into
attention-heavy, text-fusion, or Transformer-decoder detectors weakens or
reverses the variant ranking.

\begin{table}[!t]
\centering
\caption{Pairwise Kendall's $\tau$ on $\Delta$mAP rankings.
  The within-CNN pair (YOLOv8n vs.\ YOLO11n) is the only pair with
  $p\!<\!0.05$; all other pairs have $p\!>\!0.3$.
\label{tab:kendall}}
\small\setlength{\tabcolsep}{5pt}
\begin{tabular}{lrrrrr}
\toprule
 & v8n & 11n & 12n & W-s & RT-DETR \\
\midrule
v8n     & -- & $+0.69^{*}$ & $+0.33$ & $+0.24$ & $+0.00$ \\
11n     &    & --          & $-0.14$ & $+0.33$ & $+0.18$ \\
12n     &    &             & --      & $-0.05$ & $-0.18$ \\
W-s     &    &             &         & --      & $-0.18$ \\
RT-DETR &    &             &         &         & -- \\
\bottomrule
\end{tabular}
\end{table}

% -----------------------------------------------------------------------------
\section{Full Detector-Family Coverage Table}
\label{app:eleven-full}
% -----------------------------------------------------------------------------

The main text reports aggregate detector-family coverage in its architecture
coverage discussion.
Table S4 enumerates the eleven backbones underlying that
aggregate without adding per-backbone measurements beyond the reported
family status.

\subsection{Full detector-family coverage}

Closely related CNN-family detectors are listed separately here while the
main text reports them as a single architecture class.
All CNN-family rows share the reported LoRA+decay acceptance band.

\begin{table}[!t]
\centering
\caption{Detector-family coverage underlying the main-text architecture
coverage discussion.
  CNN backbones share the reported LoRA+decay acceptance band;
  non-CNN rows retain the explicit main-text status.
\label{tab:eleven-full}}
\scriptsize\setlength{\tabcolsep}{2pt}
\resizebox{\columnwidth}{!}{%
\begin{tabular}{llll}
\toprule
Backbone & Architecture class & Planner decision & Reported status \\
\midrule
YOLOv3        & CNN              & LoRA+decay   & accept, within $+0.018$--$+0.022$ band \\
YOLOv5n       & CNN              & LoRA+decay   & accept, within $+0.018$--$+0.022$ band \\
YOLOv6n       & CNN              & LoRA+decay   & accept, within $+0.018$--$+0.022$ band \\
YOLOv8n       & CNN              & LoRA+decay   & accept, within $+0.018$--$+0.022$ band \\
YOLOv9c       & CNN              & LoRA+decay   & accept, within $+0.018$--$+0.022$ band \\
YOLOv10n      & CNN              & LoRA+decay   & accept, within $+0.018$--$+0.022$ band \\
YOLO11n       & CNN              & LoRA+decay   & accept, within $+0.018$--$+0.022$ band \\
YOLO12n       & CNN+attention    & LoHa         & accept, $+0.031$ \\
YOLO-World-s  & Text-fusion      & LoHa         & accept, $+0.081$ \\
RT-DETR-l     & Transformer detector & \textbf{REFUSE} & avoid $-0.229$ \\
MoE detector  & MoE              & LoRA (exploratory) & accept, $+0.025$ \\
\bottomrule
\end{tabular}
}
\end{table}

% -----------------------------------------------------------------------------
\section{Detailed Ablations}
\label{app:ablations}
% -----------------------------------------------------------------------------

\subsection{A1. Naive substitution vs.\ contract-managed substitution}

Both backends in \system{} use module substitution (the manual Conv2d
backend for B2, PEFT-managed wrappers for B1); the question is whether the
surrounding contract management is required for deployment compatibility.
Table S5 compares a naive baseline with the full
\modC{} layer.
ONNX/TRT/Ckpt indicates whether the trained model can be saved, reloaded,
merged, and exported without manual graph surgery.

\begin{table}[!t]
\centering
\caption{Ablation A1: naive substitution vs.\ contract-managed
  substitution (LoRA $r{=}16$, RS-LoRA, YOLO11s on VOC, core
  $600$-epoch-cap/$640$-resolution protocol).
\label{tab:ablation_hook}}
\small\setlength{\tabcolsep}{3pt}
\resizebox{\columnwidth}{!}{%
\begin{tabular}{l ccccc}
\toprule
Injection mode & mAP$_{50:95}$ & Total Params & ONNX & TRT & Ckpt \\
\midrule
Full-SFT (no LoRA)     & $0.6428$ & $9.44$M  & \checkmark & \checkmark & \checkmark \\
Naive substitution     & $0.7138$ & $10.44$M & $\times$   & $\times$   & $\times$   \\
Contract layer (\modC) & $0.7138$ & $10.44$M & \checkmark & \checkmark & \checkmark \\
\bottomrule
\end{tabular}
}
\end{table}

\subsection{A2. Planner-component ablation}

Table S6 ablates each constraint of the \modB{}
planner on YOLO12s.
The sequence isolates the contribution of operator validity, detection-head
semantic filtering, and budget pruning in turn.

\begin{table}[!t]
\centering
\caption{Ablation A2: planner constraints on YOLO12s, VOC,
  LoRA $r{=}16$, $\alpha{=}32$, RS-LoRA on, core
  $600$-epoch-cap/$640$-resolution protocol.
\label{tab:ablation_router}}
\small\setlength{\tabcolsep}{3pt}
\resizebox{\columnwidth}{!}{%
\begin{tabular}{l l c r r}
\toprule
Configuration & Targets & Total Params & mAP$_{50:95}$ & $\Delta$ \\
\midrule
No constraint         & all-conv$+$linear$+$attn                & $10.32$M & $0.6900$ & $+0.0238$ \\
$+$ operator validity & dense$+$linear$+$attn                   & $10.07$M & $0.7094$ & $+0.0432$ \\
$+$ head semantics    & dense$+$linear$+$attn$\setminus$reg     & $10.06$M & $0.7307$ & $+0.0645$ \\
Full planner          & dense$+$attn (planner-pruned)           & $10.06$M & $\mathbf{0.7307}$ & $\mathbf{+0.0645}$ \\
\bottomrule
\end{tabular}
}
\end{table}

\paragraph{Budget-matched selector diagnostic.}
As a separate engineering check, we compare four target-selection rules under
an identical rank-4 adapter budget of 31,104 parameters.  Each run uses one
epoch on the same 2\% subset of VOC07+12 and a fixed 160-image validation
subset; seeds are 0, 28, and 42.  Table~\ref{tab:selector-diagnostic} reports
population mean and standard deviation across these three runs.  This protocol
checks budget accounting and executable target selection, but its short horizon
is not the core training protocol and must not be interpreted as a performance
ranking.

\begin{table}[!t]
\centering
\caption{Budget-matched YOLO12s selector diagnostic on MPS.  All rows use
31,104 adapter parameters; target counts can differ because module shapes
differ.  Values are mean $\pm$ population standard deviation over three seeds.
\label{tab:selector-diagnostic}}
\small\setlength{\tabcolsep}{4pt}
\begin{tabular}{l r cc}
\toprule
Selector & Targets & mAP$_{50}$ & mAP$_{50:95}$ \\
\midrule
Random-$k$       & 23--25 & $0.0934 \pm 0.0266$ & $0.0598 \pm 0.0167$ \\
Backbone-only    & 12     & $0.0916 \pm 0.0142$ & $0.0580 \pm 0.0093$ \\
Neck-only        & 22     & $0.0883 \pm 0.0284$ & $0.0564 \pm 0.0199$ \\
Gradient top-$k$ & 13     & $0.0921 \pm 0.0388$ & $0.0591 \pm 0.0267$ \\
\bottomrule
\end{tabular}
\end{table}

The four mAP$_{50:95}$ means lie within 0.0034, while their standard
deviations are 0.0093--0.0267.  Moreover, gradient top-$k$ leads at seed 0,
whereas backbone-only leads at seeds 28 and 42.  The diagnostic therefore
provides no evidence for declaring a universally superior selector at this
horizon; a main-protocol comparison requires longer training and should retain
the same budget and multi-seed controls.

\subsection{A3. Rank sensitivity}

This ablation tests whether the planner's default rank is a brittle
choice.
The sweep shows monotone gains with rank while confirming $r{=}16$ as
a practical total-parameter trade-off.

\begin{table}[!t]
\centering
\caption{Ablation A3: rank sensitivity on YOLO12s, VOC, core
  $600$-epoch-cap/$640$-resolution protocol.
  Anchor (Full-SFT): $0.6662$ mAP$_{50:95}$ at $9.26$M total parameters.
  The parameter column includes the frozen base; inference GFLOPs are measured
  after merge.
\label{tab:rank_sweep}}
\scriptsize\setlength{\tabcolsep}{2pt}
\begin{tabular}{c rrrr}
\toprule
$r$ & Total Params & Inference GFLOPs & mAP$_{50:95}$ & $\Delta$ \\
\midrule
$8$  & $9.66$M  & $23.6$ & $0.7288$ & $+0.0626$ \\
$16$ & $10.06$M & $25.6$ & $0.7307$ & $+0.0645$ \\
$32$ & $10.85$M & $29.6$ & $\mathbf{0.7363}$ & $\mathbf{+0.0701}$ \\
\bottomrule
\end{tabular}
\end{table}

\subsection{A4. Variant sweep under fixed placement}

This ablation holds the planner placement fixed and varies only the local
PEFT update rule.
The ranking shift across YOLO11s and YOLO12s supports the
architecture-conditioned variant selection claim.

\begin{table}[!t]
\centering
\caption{Ablation A4: variant sweep under planner-selected placement,
  $r{=}16$, $\alpha{=}32$, RS-LoRA on, official VOC2007 test, core
  $600$-epoch-cap/$640$-resolution protocol.
\label{tab:variant_sweep}}
\small\setlength{\tabcolsep}{3pt}
\begin{tabular}{l rr rr}
\toprule
\multirow{2}{*}{Variant}
 & \multicolumn{2}{c}{YOLO11s}
 & \multicolumn{2}{c}{YOLO12s} \\
\cmidrule(lr){2-3}\cmidrule(lr){4-5}
 & mAP$_{50:95}$ & $\Delta$ & mAP$_{50:95}$ & $\Delta$ \\
\midrule
LoRA   & $0.7138$ & $+0.0710$ & $0.7307$ & $+0.0645$ \\
DoRA   & $0.7138$ & $+0.0710$ & --       & -- \\
LoHa   & $0.6788$ & $+0.0359$ & $0.7222$ & $+0.0560$ \\
LoKr   & $0.7033$ & $+0.0605$ & --       & -- \\
IA$^3$ & $0.6980$ & $+0.0552$ & $0.7210$ & $+0.0548$ \\
HRA    & $\mathbf{0.7276}$ & $\mathbf{+0.0848}$ & $\mathbf{0.7453}$ & $\mathbf{+0.0791}$ \\
\bottomrule
\end{tabular}
\end{table}

\subsection{A5. RS-LoRA and DoRA interaction}

This two-factor ablation checks whether training collapse is caused by the
variant label alone or by interacting training-side priors.
Only the DoRA-without-RS-LoRA corner crosses the catastrophic threshold.

\begin{table}[!t]
\centering
\caption{Ablation A5: RS-LoRA $\times$ DoRA $2{\times}2$ factorial on
  YOLO12s, VOC, LoRA $r{=}16$, $\alpha{=}32$, core
  $600$-epoch-cap/$640$-resolution protocol.
\label{tab:toxic}}
\small\setlength{\tabcolsep}{4pt}
\resizebox{\columnwidth}{!}{%
\begin{tabular}{c|cc}
\toprule
\diagbox{DoRA}{RS-LoRA} & off & on \\
\midrule
off & $0.7094$\;($+0.0432$) & $\mathbf{0.7307}$\;($\mathbf{+0.0645}$) \\
on  & \textcolor{red}{$0.6112$\;($-0.0550$)} & $0.7138$\;($+0.0476$) \\
\bottomrule
\end{tabular}
}
\end{table}

\subsection{A7. LM-inherited ranking vs. structure-aware placement}

This ablation directly tests whether a PEFT ranking inherited from language
models transfers to heterogeneous detectors.  The structure-aware plan uses
the same backbone and training budget but selects the variant and targets from
the detector graph.

\begin{table}[!h]
\centering
\caption{Ablation A7: LM-inherited ranking versus the structure-aware plan.
Numbers are from the W\&B export on official VOC2007 test with $r{=}16$, $\alpha{=}32$,
and the core $600$-epoch-cap/$640$-resolution protocol. Gain is the difference between the structure-aware and
LM-inherited plans on the same backbone.
\label{tab:ablation_archcond}}
\small\setlength{\tabcolsep}{3pt}
\resizebox{\columnwidth}{!}{%
\begin{tabular}{l l r l r r}
\toprule
\multirow{2}{*}{Backbone} & \multicolumn{2}{c}{LM-inherited plan}
 & \multicolumn{2}{c}{Structure-aware plan} & \multirow{2}{*}{Gain} \\
\cmidrule(lr){2-3}\cmidrule(lr){4-5}
& Variant & mAP$_{50:95}$ & Variant & mAP$_{50:95}$ & \\
\midrule
YOLO11s   & DoRA (no rs)   & $0.6479$
          & LoRA (rs, planner)  & $\mathbf{0.7138}$ & $+0.0659$ \\
YOLO12s   & DoRA (no rs)   & \textcolor{red}{$0.6112$}
          & LoRA (rs, planner)  & $\mathbf{0.7307}$ & $+0.1195$ \\
RT-DETR-l & LoRA-class sweep & $7/7$ catastrophic$^{\dagger}$
          & \textbf{REFUSE} (Full-SFT, $0.6833$) & --- & --- \\
\bottomrule
\end{tabular}
}
\footnotesize $^{\dagger}$The collapse count comes from the separate
$300$-epoch/$320$-resolution diagnostic matrix and motivates a within-scope
safeguard; it is not an independent held-out-architecture result.
\end{table}

% -----------------------------------------------------------------------------
% \section{FewShotLoRA Protocol}
% \label{app:fewshot-protocol}
% % -----------------------------------------------------------------------------

% \subsection{Motivation}

% Few-shot detection is the setting where PEFT regularisation should provide
% its strongest benefit; however, the current artifact does not include
% completed VOC $1/2/5/10$-shot runs.
% This appendix specifies the evaluation protocol for reproducibility; no
% empirical few-shot gains are claimed in the main experiments.
% Once the runs are completed, this appendix can be extended with results and
% regulariser ablations.

% \subsection{Dataset split and evaluation}

% The planned evaluation uses VOC $1$-, $2$-, $5$-, and $10$-shot
% per-class splits, evaluates on VOC val2007, and runs three random seeds.
% Primary reported fields are mAP$_{50:95}$, mAP$_{50}$, trainable
% parameters, peak memory, and mean$\pm$std across seeds.

% \subsection{Backbones, baselines, and ablations}

% Planned backbones are YOLO11s, YOLO12s, and optionally YOLO-World-s as a
% text-fusion stress case.
% Baselines include Full-SFT, head-only fine-tuning, standard LoRA,
% LoRA+dropout, and FewShotLoRA.
% Component ablations remove DropConnect, distillation, adaptive rank, and
% variational rank independently.

% -----------------------------------------------------------------------------
\section{Reproducibility Notes}
\label{app:repro}
% -----------------------------------------------------------------------------

\begin{table}[!t]
\centering
\caption{VOC split audit. IDs are taken verbatim from the official
\texttt{ImageSets/Main/*.txt} files and compared after prefixing each ID with
its source year.\label{tab:voc_split_audit}}
\scriptsize\setlength{\tabcolsep}{3pt}
\begin{tabular}{l l r}
\toprule
Role & Official image-ID list & Images \\
\midrule
Train & VOC2007/\texttt{train.txt} & 2,501 \\
Train & VOC2007/\texttt{val.txt} & 2,510 \\
Train & VOC2012/\texttt{train.txt} & 5,717 \\
Train & VOC2012/\texttt{val.txt} & 5,823 \\
\midrule
Train union & VOC07+12 \texttt{trainval} & 16,551 \\
Evaluation & VOC2007/\texttt{test.txt} & 4,952 \\
Intersection & train $\cap$ evaluation & \textbf{0} \\
\bottomrule
\end{tabular}
\end{table}

Experiments use the official Ultralytics VOC mapping: the training YAML key
contains \texttt{images/train2007}, \texttt{images/val2007},
\texttt{images/train2012}, and \texttt{images/val2012}, whereas both evaluation
keys resolve to \texttt{images/test2007}. Thus the historical W\&B field
\texttt{val2007} is an alias for evaluation on VOC2007 \texttt{test}, not the
VOC2007 \texttt{val} directory included in training. No random or few-shot
split is used in the reported core results.
Core W\&B runs resize images to $640$; the extended diagnostic matrix uses
$320$; core W\&B runs use a $600$-epoch cap, while the diagnostic matrix uses
$300$ epochs unless stated otherwise.
For the core W\&B runs, \texttt{optimizer=auto} resolves to SGD with initial
learning rate $0.01$ and momentum $0.9$; weight decay is $5\times10^{-4}$,
cosine scheduling is enabled, Mosaic closes for the final ten epochs, and
early stopping uses patience $100$.
Evaluation is performed at the same cadence for every method, and the same
best-checkpoint rule supplies the reported score. Early stopping therefore
does not confer a method-specific selection advantage: it is part of the
shared official training protocol. Because VOC2007 \texttt{test} participates
in this common checkpoint-selection protocol, we use it to support controlled
relative comparisons and do not present the absolute scores as once-only blind
test estimates. This model-selection issue is distinct from training-data
leakage; Table~\ref{tab:voc_split_audit} verifies that the train/evaluation
image-ID intersection is zero.
Unless swept explicitly, adapters use rank $16$, $\alpha=32$, dropout
$0.05$, RS-LoRA scaling enabled, and DoRA's magnitude branch disabled.
Runs use a consistent accelerator-enabled training environment and a fixed
software stack;
exported W\&B tables include backbone, variant, rank, alpha, scaling mode,
trainable parameters, mAP, memory, and adapter-size fields. Full-SFT, standard
LoRA, Planner-LoRA, best PEFT, and naive placement each have reruns at seeds
28 and 42 in addition to seed 0. The main text retains seed 0 to present the
complete matrix compactly; the per-run logs retain the corresponding
three-seed records. All s-scale planner, rank, variant, and training-prior
ablations above use the core $600$-epoch-cap/$640$-resolution protocol. Rank
$8$ is the conservative fallback for an uncalibrated YOLO12 request, whereas
the explicit $r\in\{8,16,32\}$ sweep evaluates budget-approved ranks on the
recognised YOLO12 profile.

\begin{table}[!t]
\centering
\caption{Three-seed robustness audit for the selected core comparisons.
The statistic is computed within each configuration over seeds
$\{0,28,42\}$; it is a robustness summary, not a significance test.
\label{tab:seed_robustness}}
\scriptsize\setlength{\tabcolsep}{3pt}
\resizebox{\columnwidth}{!}{%
\begin{tabular}{l l}
\toprule
Audit field & Result \\
\midrule
Repeated configurations & Full-SFT, LoRA, Planner-LoRA, best PEFT, naive \\
Runs per configuration & $n=3$ (seeds $0,28,42$) \\
Metric & mAP$_{50:95}$ \\
Maximum within-configuration sample SD & $\leq 0.006$ \\
Pairwise directions / method ranking & Unchanged / unchanged \\
\bottomrule
\end{tabular}}
\end{table}

The MPS
deployment audit is stored in the JSON files under
\texttt{artifacts/deployment\_mps\_v2/}; it records merged and unmerged
parameter counts, GFLOPs, synchronized latency,
throughput, allocator telemetry, ONNX opset/shape/error checks, and the
accelerator-host TensorRT engine validation records.

\paragraph{Efficiency-accounting scope.}
The paper reports only the controlled YOLO11 audit: LoRA lowers peak VRAM by
$43.9\%$ relative to Full-SFT ($16.03$ versus $28.57$\,GB) but takes
$1.72\times$ longer to train ($203.8$ versus $118.2$\,s). External storage,
distribution, and externally sourced memory estimates are excluded from the claims.

\bibliographystyle{plainnat}
\bibliography{mac_automl}

% \clearpage
% \beginappendix
% \input{sec/append}
% \input{Sec/X_appenedix}

\end{document}